\ifdefined\pdfoutput\pdfoutput=1\fi
\documentclass[10pt,twocolumn,letterpaper]{article}
\usepackage[T1]{fontenc}
\usepackage[pagenumbers]{sadstyle}

\usepackage{cuted} 

\usepackage{currfile} 

\usepackage{caption} 

\usepackage{booktabs}
\usepackage{multirow}
\usepackage{amsmath,amssymb}
\usepackage{xspace}
\usepackage{tikz}
\usetikzlibrary{arrows.meta,positioning,fit,backgrounds,calc}
\newcommand{\method}{SAD\xspace}
\newcommand{\zgen}{z_{\text{gen}}}

\newcommand{\Ldist}{\mathcal{L}_{\text{GSA}}}
\newcommand{\Lanc}{\mathcal{L}_{\text{EAP}}}
\newcommand{\fdclip}{FD$_\text{CLIP}$\xspace}
\newcommand{\fddino}{FD$_\text{DINO}$\xspace}
\newcommand{\gain}[1]{}
\newcommand{\loss}[1]{}
\definecolor{ForestGreen}{rgb}{0.13,0.55,0.13}
\definecolor{BrickRed}{rgb}{0.70,0.13,0.13}
\usepackage{colortbl}
\definecolor{oursbg}{RGB}{232,241,250}
\definecolor{gaincol}{RGB}{30,130,60}
\definecolor{losscol}{RGB}{190,40,40}
\newcommand{\hdr}[1]{\multicolumn{1}{c}{#1}}

\usetikzlibrary{shapes.geometric,shapes.misc,arrows.meta,positioning,calc,fit,backgrounds,decorations.pathmorphing}

\definecolor{frzfill}{RGB}{236,190,222}   
\definecolor{frzdraw}{RGB}{196,120,176}
\definecolor{trnfill}{RGB}{160,212,240}   
\definecolor{trndraw}{RGB}{58,140,196}
\definecolor{objfill}{RGB}{252,232,160}   
\definecolor{objdraw}{RGB}{214,170,40}
\definecolor{distband}{RGB}{232,242,252}  
\definecolor{ancband}{RGB}{233,246,233}   
\definecolor{gradred}{RGB}{214,60,60}
\definecolor{gradgreen}{RGB}{46,150,70}
\definecolor{tokfake}{RGB}{247,170,120}
\definecolor{tokreal}{RGB}{140,200,140}
\definecolor{boxgray}{RGB}{226,229,233}
\definecolor{txtgray}{RGB}{100,100,100}

\tikzset{
  arr/.style={-{Stealth[length=1.6mm,width=1.4mm]}, line width=0.6pt, black!75},
  garr/.style={-{Stealth[length=1.8mm,width=1.6mm]}, line width=0.9pt, densely dashed},
  lbl/.style={font=\scriptsize, inner sep=1pt},
  var/.style={font=\small, inner sep=1pt},
  graybox/.style={draw=black!25, fill=boxgray, rounded corners=3pt, font=\scriptsize, align=center, inner sep=3pt, minimum height=7mm},
  obj/.style={draw=objdraw, fill=objfill, rounded corners=4pt, line width=0.6pt, align=center, inner sep=3pt},
  pcap/.style={font=\footnotesize\bfseries, text=black!80},
}

\newcommand{\trap}[8]{
  \coordinate (#1) at #2;
  \filldraw[fill=#6, draw=#7, line width=0.6pt, line join=round]
    ($(#1)+(-#3/2,#4/2)$) -- ($(#1)+(#3/2,#5/2)$) -- ($(#1)+(#3/2,-#5/2)$) -- ($(#1)+(-#3/2,-#4/2)$) -- cycle;
  \node[font=\scriptsize, inner sep=0pt] at (#1) {#8};
  \coordinate (#1-w) at ($(#1)+(-#3/2,0)$); \coordinate (#1-e) at ($(#1)+(#3/2,0)$);
  \coordinate (#1-n) at ($(#1)+(0,{max(#4,#5)/2})$); \coordinate (#1-s) at ($(#1)+(0,{-max(#4,#5)/2})$);
  \coordinate (#1-ne) at ($(#1)+(#3/2,{#5/2})$); \coordinate (#1-nw) at ($(#1)+(-#3/2,{#4/2})$);
}

\definecolor{tokenc}{RGB}{190,221,245}    
\definecolor{tokgen}{RGB}{249,201,160}    
\definecolor{tokrf}{RGB}{198,232,193}     
\definecolor{tokgf}{RGB}{246,183,176}     
\definecolor{barpeach}{RGB}{252,233,200}  
\definecolor{barlilac}{RGB}{228,224,246}  
\definecolor{lossbox}{RGB}{251,221,214}   
\definecolor{lossdraw}{RGB}{220,140,120}
\definecolor{decfill}{RGB}{190,221,245}
\definecolor{decdraw}{RGB}{110,165,210}

\newcommand{\vbar}[6]{%
  \node[fill=#5, rounded corners=4pt, minimum width=#3, minimum height=#4, inner sep=0pt, font=\scriptsize\bfseries, align=center] (#1) at #2 {#6};}
\newcommand{\tokgrid}[6]{%
  \pgfmathsetmacro{\tgw}{#3*#6+(#3-1)*0.05}\pgfmathsetmacro{\tgh}{#4*#6+(#4-1)*0.05}
  \begin{scope}[shift={($#2+(-\tgw/2,-\tgh/2)$)}]
    \foreach \ii in {1,...,#3} \foreach \jj in {1,...,#4} {
      \pgfmathsetmacro{\shade}{55+mod(37*\ii+53*\jj+11*\ii*\jj,45)}
      \fill[#5!\shade!white, rounded corners=0.8pt] ({(\ii-1)*(#6+0.05)},{(\jj-1)*(#6+0.05)}) rectangle ++(#6,#6);}
  \end{scope}
  \path ($#2+(-\tgw/2,-\tgh/2)$) coordinate (#1-sw) ($#2+(\tgw/2,\tgh/2)$) coordinate (#1-ne);
  \coordinate (#1-n) at ($#2+(0,\tgh/2)$); \coordinate (#1-s) at ($#2+(0,-\tgh/2)$);
  \coordinate (#1-w) at ($#2+(-\tgw/2,0)$); \coordinate (#1-e) at ($#2+(\tgw/2,0)$);
}
\newcommand{\snow}[1]{%
  \begin{scope}[shift={#1}, line width=0.55pt, draw=cyan!60!blue!70, line cap=round]
    \foreach \a in {0,60,120} {\draw ({0.11*cos(\a)},{0.11*sin(\a)}) -- ({-0.11*cos(\a)},{-0.11*sin(\a)});}
    \foreach \a in {0,60,...,300} {\draw ({0.07*cos(\a)},{0.07*sin(\a)}) -- ++({0.035*cos(\a+40)},{0.035*sin(\a+40)});
                                    \draw ({0.07*cos(\a)},{0.07*sin(\a)}) -- ++({0.035*cos(\a-40)},{0.035*sin(\a-40)});}
  \end{scope}}
\newcommand{\fire}[1]{%
  \begin{scope}[shift={#1}, scale=0.13]
    \fill[red!75!orange] (0,-1) .. controls (-0.9,-1) and (-1,0) .. (-0.55,0.55) .. controls (-0.5,0.2) and (-0.25,0.15) .. (-0.2,0.35)
      .. controls (-0.3,0.9) and (0.1,1.2) .. (0.15,1.35) .. controls (0.2,0.9) and (1.0,0.6) .. (0.75,-0.2) .. controls (0.65,-0.75) and (0.4,-1) .. (0,-1);
    \fill[orange!60!yellow] (0,-0.85) .. controls (-0.45,-0.85) and (-0.5,-0.2) .. (-0.2,0.1) .. controls (-0.1,-0.1) and (0.1,0.1) .. (0.05,0.45)
      .. controls (0.35,0.1) and (0.5,-0.3) .. (0.35,-0.6) .. controls (0.25,-0.8) and (0.1,-0.85) .. (0,-0.85);
  \end{scope}}
\tikzset{
  ptitle/.style={font=\small\bfseries},
  darr/.style={-{Stealth[length=1.6mm,width=1.4mm]}, line width=0.55pt, densely dashed, black!70},
  sep/.style={black!70, dashed, line width=1.1pt},
}
\definecolor{trainfill}{RGB}{255,226,140}
\definecolor{traindraw}{RGB}{214,158,24}
\definecolor{genfill}{RGB}{224,229,238}

\definecolor{pAdraw}{RGB}{84,141,212}  \definecolor{pAfill}{RGB}{240,246,253}   
\definecolor{pBdraw}{RGB}{106,168,79}  \definecolor{pBfill}{RGB}{243,249,239}   
\definecolor{pCdraw}{RGB}{230,162,40}  \definecolor{pCfill}{RGB}{254,248,234}   
\definecolor{ptag}{RGB}{0,0,0}
\newcommand{\panel}[5]{%
  \filldraw[draw=#3, fill=#4, line width=0.8pt, densely dashed, rounded corners=5pt] #1 rectangle #2;
  \path #1 -- #2 coordinate[pos=0] (pp-a) coordinate[pos=1] (pp-b);
  \node[anchor=north west, font=\scriptsize\bfseries\itshape, text=ptag, inner sep=2pt, xshift=4pt, yshift=-3pt] at (pp-a |- pp-b) {#5};}

\definecolor{gborder}{RGB}{214,150,70}\definecolor{pborder}{RGB}{140,128,200}  

\usepackage{pifont}
\usepackage{pgfplots}
\pgfplotsset{compat=1.17}
\usetikzlibrary{plotmarks}
\definecolor{cDz}{RGB}{88,120,160}
\definecolor{cSAD}{RGB}{226,138,18}
\definecolor{cEnc}{RGB}{198,72,72}
\definecolor{cNoise}{RGB}{140,140,140}
\definecolor{cPair}{RGB}{38,150,140}
\definecolor{cRef}{RGB}{136,96,180}
\definecolor{cGood}{RGB}{232,245,232}
\definecolor{cGrid}{RGB}{226,226,226}
\pgfplotsset{
  sadplot/.style={
    width=\linewidth, height=0.62\linewidth,
    axis lines=left, axis line style={-, black!60, line width=0.4pt},
    tick style={black!60, line width=0.3pt}, tick align=outside, tick pos=left,
    major tick length=2pt,
    grid=major, grid style={cGrid!70, line width=0.25pt},
    font=\scriptsize, label style={font=\scriptsize}, tick label style={font=\scriptsize},
    legend style={font=\tiny, draw=none, fill=white, fill opacity=0.85, text opacity=1, inner sep=1pt,
                  row sep=-1pt, legend cell align=left},
    every axis plot/.append style={line width=0.9pt},
    title style={font=\scriptsize\bfseries, yshift=-4pt},
    clip=false,
  },
  ptD/.style={cDz, mark=*, mark size=1.5pt, mark options={fill=cDz}},
  ptS/.style={cSAD, mark=*, mark size=1.6pt, mark options={fill=cSAD}},
  ptE/.style={cEnc, mark=triangle*, mark size=1.9pt, mark options={fill=cEnc}},
}
\pgfplotsset{discard if not/.style 2 args={x filter/.append code={%
  \edef\tempa{\thisrow{#1}}\edef\tempb{#2}\ifx\tempa\tempb\else\fi}}}
\usetikzlibrary{arrows.meta}
\usetikzlibrary{patterns}

\usepackage{amsthm}
\usepackage{algorithm}
\usepackage{algpseudocode}
\algrenewcommand\algorithmiccomment[1]{\hfill{\color{black!55}\scriptsize$\triangleright$ #1}}
\theoremstyle{plain}
\newtheorem{proposition}{Proposition}
\usepackage{needspace}
\newcommand{\simg}[2]{\includegraphics[width=#1]{#2}}

\newcommand{\figsuppunc}{%
\begin{figure*}[t]
\centering
\setlength{\tabcolsep}{1pt}
\renewcommand{\arraystretch}{0}
\def\us{2.78cm}
\begin{tabular}{@{}r@{\hspace{3pt}}cccccc@{}}
\rotatebox{90}{\hspace{0.9cm}\scriptsize $D_0$} & \simg{\us}{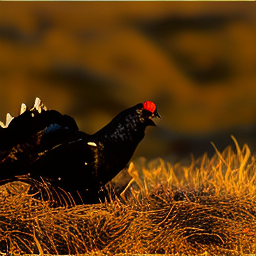} & \simg{\us}{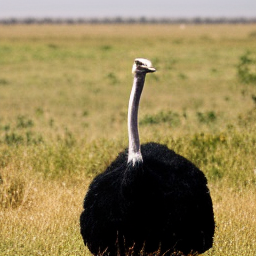} & \simg{\us}{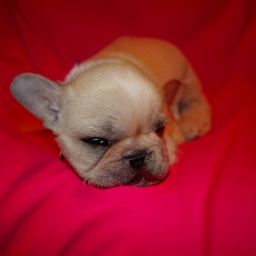} & \simg{\us}{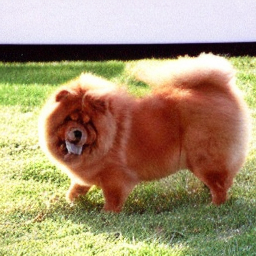} & \simg{\us}{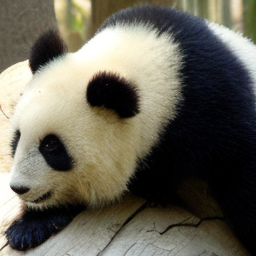} & \simg{\us}{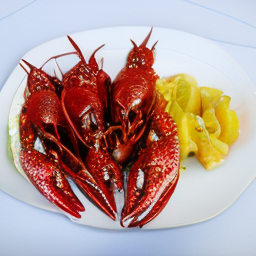} \\[1pt]
\rotatebox{90}{\hspace{0.8cm}\scriptsize \method} & \simg{\us}{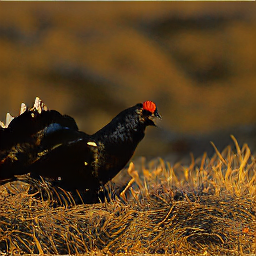} & \simg{\us}{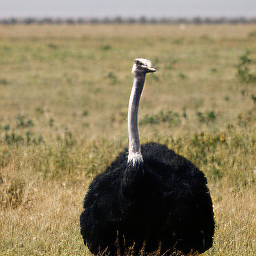} & \simg{\us}{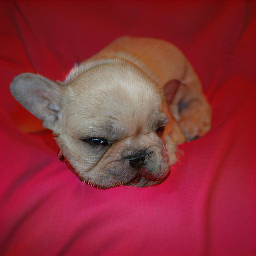} & \simg{\us}{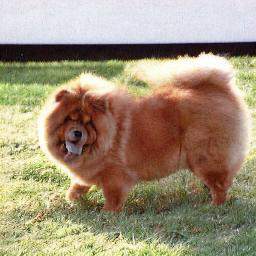} & \simg{\us}{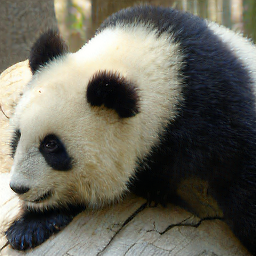} & \simg{\us}{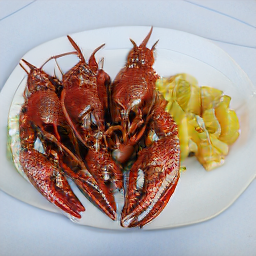} \\
\rotatebox{90}{\makebox[\us][c]{\scriptsize $|\text{\method}-D_0|$}} & \simg{\us}{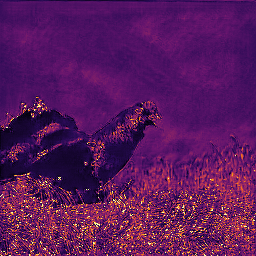} & \simg{\us}{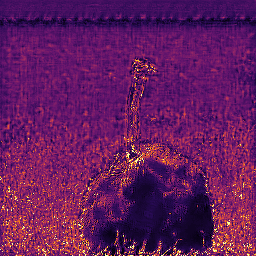} & \simg{\us}{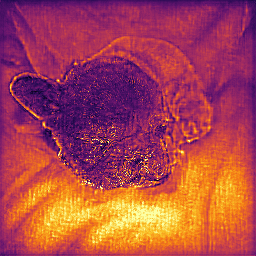} & \simg{\us}{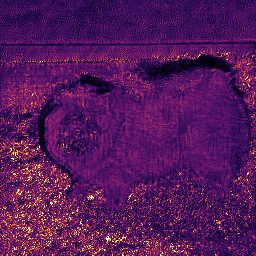} & \simg{\us}{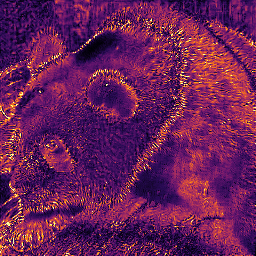} & \simg{\us}{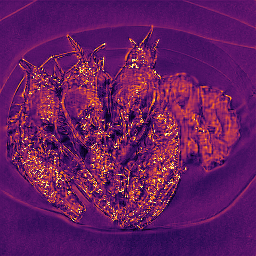} \\[5pt]
\rotatebox{90}{\hspace{0.9cm}\scriptsize $D_0$} & \simg{\us}{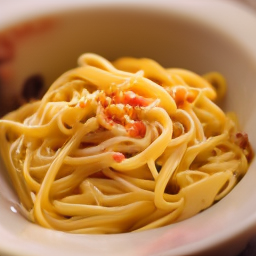} & \simg{\us}{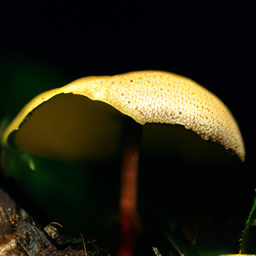} & \simg{\us}{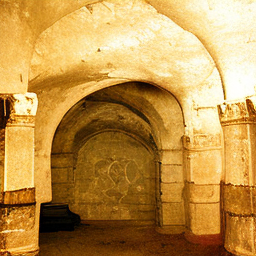} & \simg{\us}{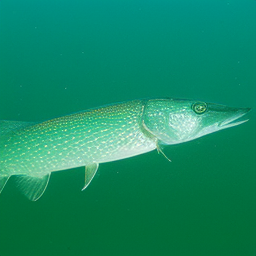} & \simg{\us}{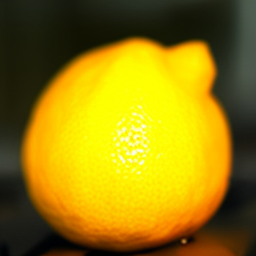} & \simg{\us}{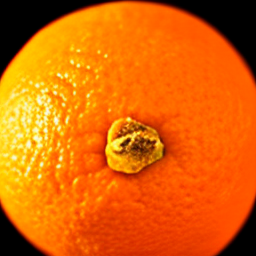} \\[1pt]
\rotatebox{90}{\hspace{0.8cm}\scriptsize \method} & \simg{\us}{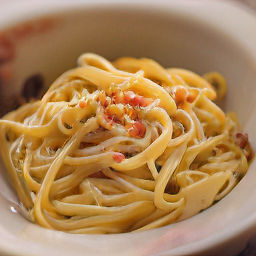} & \simg{\us}{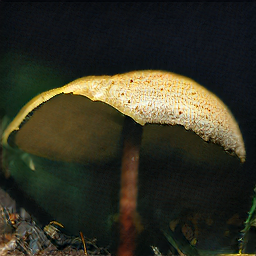} & \simg{\us}{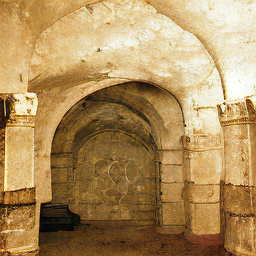} & \simg{\us}{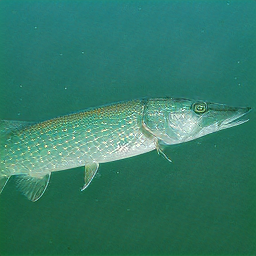} & \simg{\us}{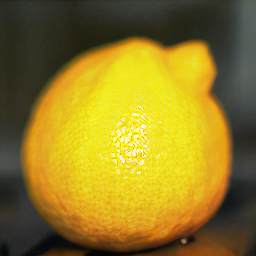} & \simg{\us}{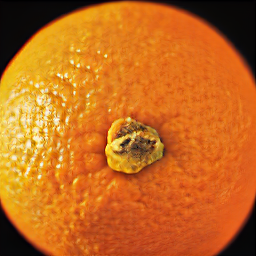} \\
\rotatebox{90}{\makebox[\us][c]{\scriptsize $|\text{\method}-D_0|$}} & \simg{\us}{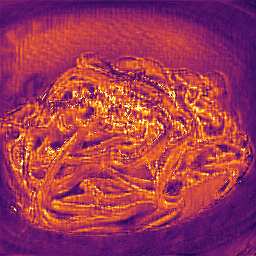} & \simg{\us}{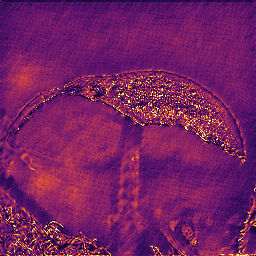} & \simg{\us}{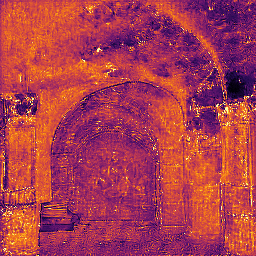} & \simg{\us}{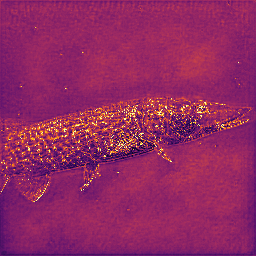} & \simg{\us}{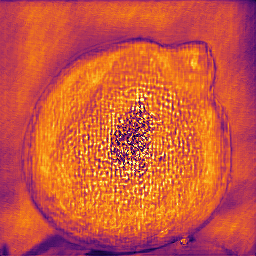} & \simg{\us}{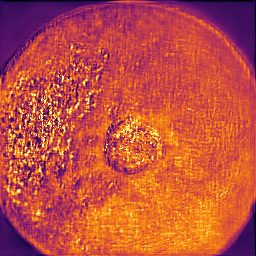} \\
\end{tabular}
\caption{\textbf{Additional samples decoded by $D_0$ and \method.} DiT-XL/2 in the ablation setting (guidance 1.5); the last two samples use guidance 4.0, which is not seen in training. The third and sixth rows show the per-pixel absolute difference (brighter is larger).}
\label{fig:supp_unc}
\end{figure*}}

\newcommand{\figsuppcoco}{%
\begin{figure*}[t]
\centering
\setlength{\tabcolsep}{0.4pt}
\renewcommand{\arraystretch}{0}
\def\cs{1.86cm}
\begin{tabular}{@{}ccc@{\hspace{6pt}}ccc@{\hspace{6pt}}ccc@{}}
{\scriptsize $D_0$} & {\scriptsize $\lambda{=}0.01$} & {\scriptsize $\lambda{=}0.05$} & {\scriptsize $D_0$} & {\scriptsize $\lambda{=}0.01$} & {\scriptsize $\lambda{=}0.05$} & {\scriptsize $D_0$} & {\scriptsize $\lambda{=}0.01$} & {\scriptsize $\lambda{=}0.05$} \\
\simg{\cs}{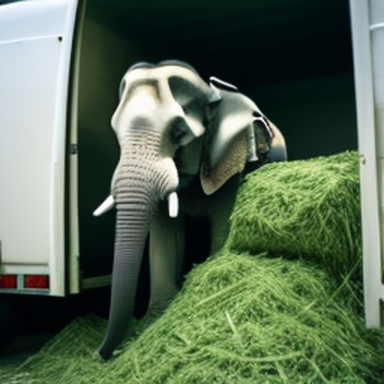} & \simg{\cs}{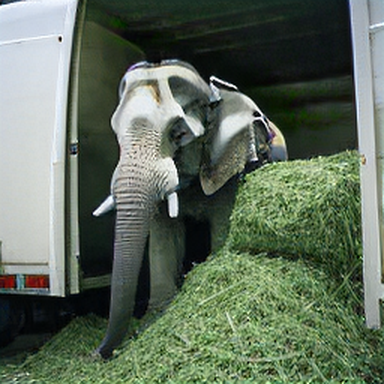} & \simg{\cs}{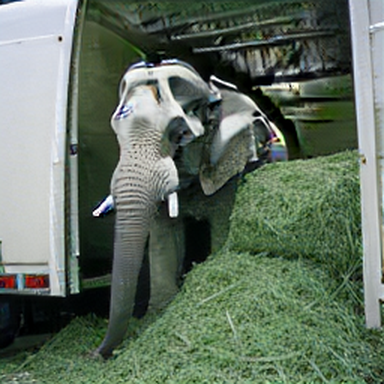} & \simg{\cs}{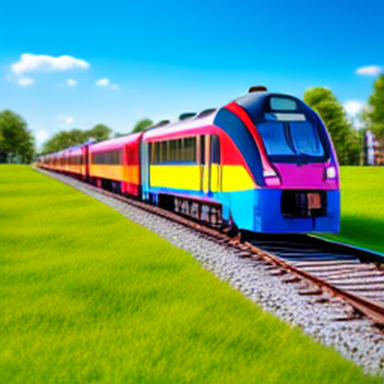} & \simg{\cs}{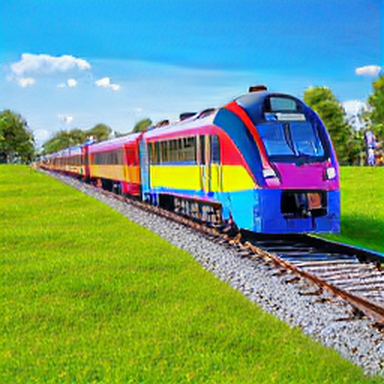} & \simg{\cs}{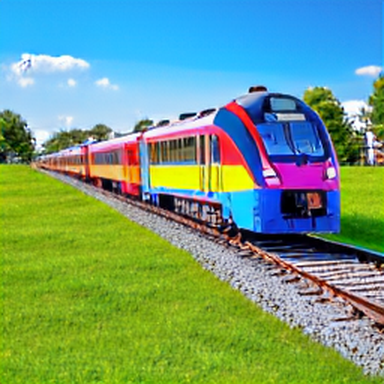} & \simg{\cs}{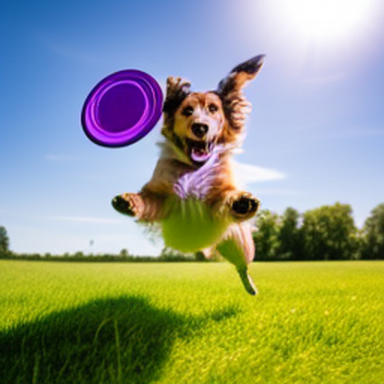} & \simg{\cs}{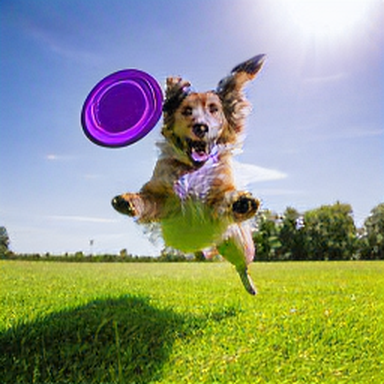} & \simg{\cs}{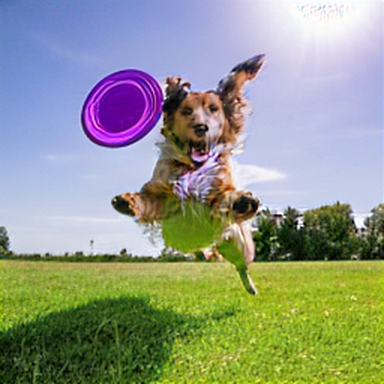} \\[1.5pt]
\simg{\cs}{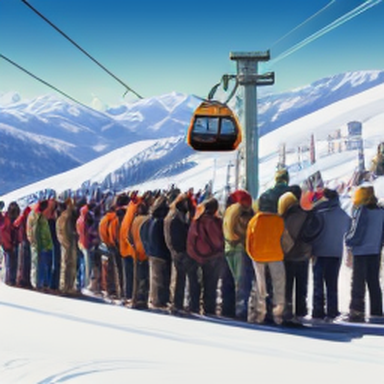} & \simg{\cs}{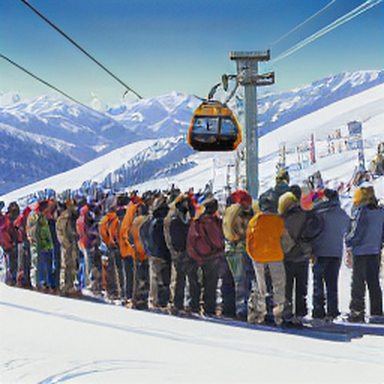} & \simg{\cs}{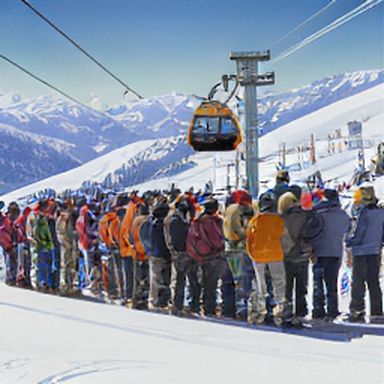} & \simg{\cs}{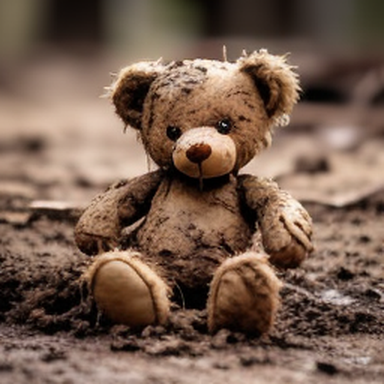} & \simg{\cs}{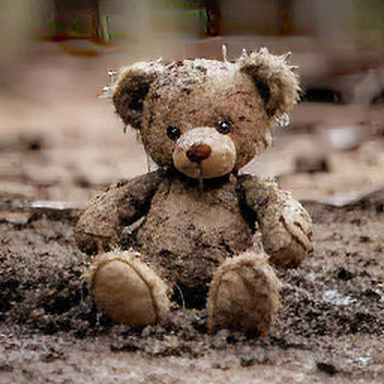} & \simg{\cs}{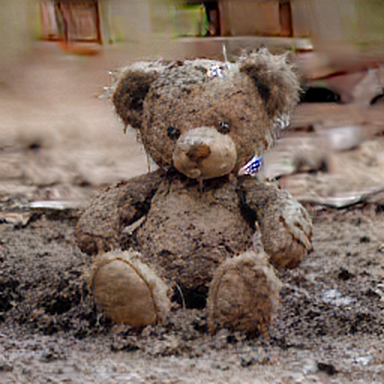} & \simg{\cs}{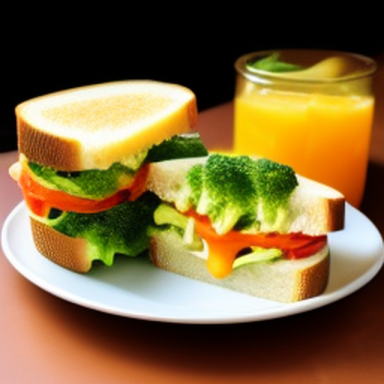} & \simg{\cs}{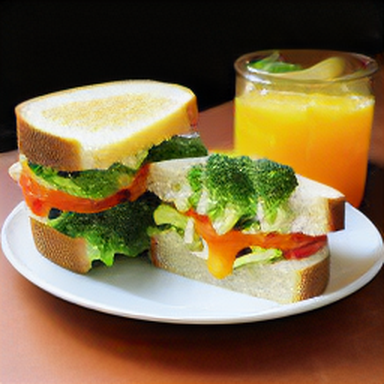} & \simg{\cs}{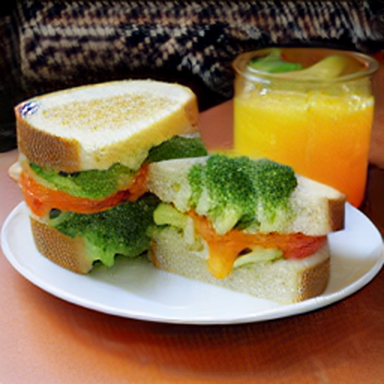} \\[1.5pt]
\simg{\cs}{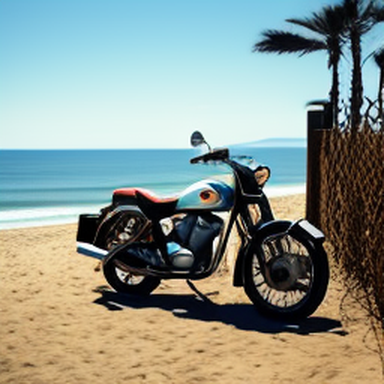} & \simg{\cs}{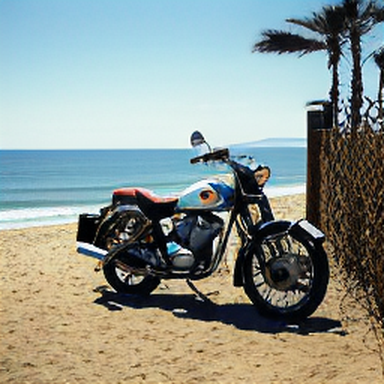} & \simg{\cs}{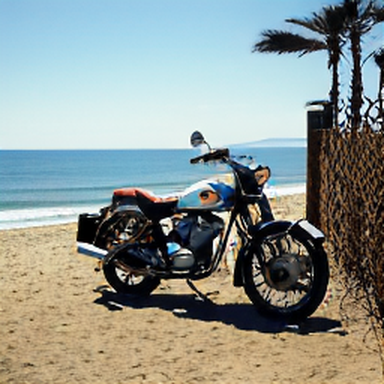} & \simg{\cs}{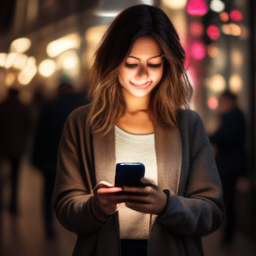} & \simg{\cs}{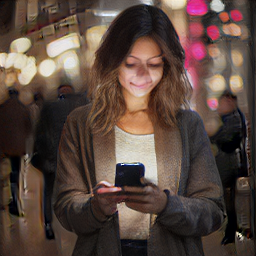} & \simg{\cs}{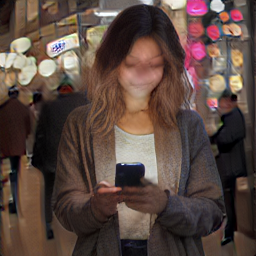} & \simg{\cs}{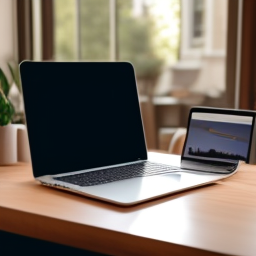} & \simg{\cs}{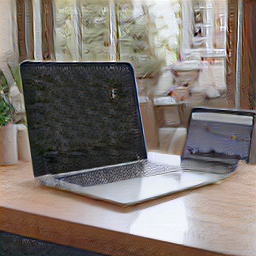} & \simg{\cs}{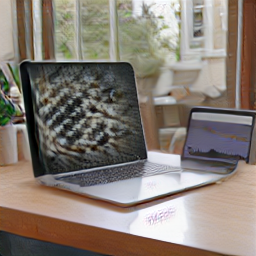} \\[1.5pt]
\end{tabular}
\caption{\textbf{PixArt-$\alpha$ samples decoded by $D_0$ and by \method-GAN at two values of $\lambda$.} The last two examples show the failure modes at the larger $\lambda$.}
\label{fig:supp_coco}
\end{figure*}}

\usepackage{makecell}

\newcommand{\phead}[1]{\par\vspace{4pt plus 1pt minus 1.5pt}\noindent\textbf{#1}}
\newcommand{\rhead}[1]{\par\vspace{5pt plus 1pt minus 1.5pt}\noindent\textbf{#1}}

\definecolor{linkblue}{rgb}{0.21,0.49,0.74}
\usepackage[pagebackref,breaklinks,colorlinks,allcolors=linkblue]{hyperref}

\title{Sample-Aligned Decoders for Latent Generative Models}

\newcommand{\pageurl}{https://torpedo2003.github.io/SAD/}
\author{%
Yu Lei$^{1,2}$\qquad Yi Li$^{2}$\qquad Hongcheng Gao$^{3}$\qquad Yipeng Wei$^{2}$\\
Liang Li$^{1}$\qquad Li Su$^{2}$\qquad Qingming Huang$^{2}$\\[0.6em]
{\normalsize $^{1}$Institute of Computing Technology, Chinese Academy of Sciences}\\[0.15em]
{\normalsize $^{2}$University of Chinese Academy of Sciences\qquad $^{3}$Tsinghua University}
}

\usepackage{pifont}
\begin{document}
\maketitle

\begin{abstract}
Latent generative models synthesize images by decoding generated latents, yet their decoder is trained only to reconstruct encoder latents and never sees the latents it decodes at test time, so the residual errors of the generator are rendered directly into the final image.
Existing remedies retrain the generator, add cost at every sampling step, or make the decoder robust to perturbed real latents, assuming that generator errors resemble such perturbations.
We find that they do not. Generated latents lie on the encoder latent manifold as closely as real ones, yet follow a different distribution.
The decoder should therefore be adapted to the distribution of generated latents, which raises two challenges: (1)~\textbf{unpaired supervision}, since a generated latent has no ground-truth image, and (2)~\textbf{latent-space drift}, since a decoder freed from paired targets forgets the latent space it must decode.
We propose \textbf{Sample-Aligned Decoders (\method)}, a post-training framework that fine-tunes only the decoder of a frozen generator at no inference cost.
\emph{Generated-Sample Distribution Alignment} matches decoded generated latents to unpaired real images in a frozen foundation-model feature space, and \emph{Encoder Anchor Preservation} keeps the decoder close to the original one on encoder latents.
Our analysis shows that the alignment is effective only on generated latents, where it corrects each sample in proportion to how much more often the generator produces such latents than the encoder; on encoder latents it can only work against the anchor.
\method improves four state-of-the-art ImageNet generators across three tokenizers, \eg, FID-50K from $1.15$ to $0.98$ for REPA-E, complements discriminator guidance, and extends to text-to-image generation. Our code and models are available at \url{\pageurl}.
\end{abstract}

\section{Introduction}
\label{sec:intro}

Latent generative models~\cite{rombach2022ldm,peebles2023dit,ma2024sit,podell2024sdxl} have become the dominant paradigm for high-resolution image synthesis.
An autoencoder (or visual tokenizer)~\cite{kingma2014vae,esser2021taming} maps images into a compact latent space, a generator is trained entirely in this space, and at sampling time the decoder turns every generated latent into the final image.
The two networks are trained in separate stages, and recent progress has come mostly from better latent spaces and better generators~\cite{yao2025vavae,yu2025repa,leng2025repae,chen2025maetok}, both of which require retraining the generator at considerable cost.

\input{fig/teaser}

The decoder, however, is trained only to reconstruct encoder latents $E(x)$, whereas at sampling time it decodes generated latents $\zgen$ (\cref{fig:teaser}a).
No training stage ever checks how generated latents look after decoding, so the residual errors of the generator, such as missing texture and unnatural colors, pass directly to the final image.

Existing remedies leave this mismatch in place.
Methods that improve the latent space or the generator~\cite{yao2025vavae,yu2025repa,leng2025repae} must retrain the generator, which costs hundreds to thousands of GPU-hours, and the new decoder is again trained only on encoder latents.
Sampling-time corrections such as guidance~\cite{ho2022cfg,karras2024autoguidance} or discriminator guidance~\cite{kim2023dg} act on every sampling step and therefore add inference cost.
Decoder-side methods make the decoder robust to random noise on the latents~\cite{zheng2025rae} or to perturbed real latents with paired targets~\cite{qiu2025robustok}, which implicitly assumes that generator errors look like such perturbations.

We find that this assumption does not hold.
Our analysis shows that re-encoding the decodings of generated latents changes them as little as it changes real latents, so generated latents are not off the encoder manifold, yet a small classifier easily separates the two.
The residual errors of the generator thus shift the distribution on the manifold instead of perturbing latents off it, and the decoder has to be adapted to the \emph{distribution} of generated latents.
Robust decoders instead spend their capacity on perturbations that the generator rarely produces, while the errors it does produce remain uncorrected.
The decoder is an attractive place for this adaptation, because it is an order of magnitude smaller than the generator, runs once per sample, and can be adapted without back-propagation through the generator.

This adaptation faces two challenges.
\textbf{(1) Unpaired supervision.} Generated latents have no ground-truth images, and supervision on encoder latents, where real images are available, does not reach the errors of the generator, so aligning decoded encoder latents with real images neither improves generation nor preserves reconstruction.
\textbf{(2) Latent-space drift.} Without paired targets, the decoder can satisfy its objective by forgetting how to decode the latent space, which degrades reconstruction and overfits the feature space used for supervision.

To address these challenges, we propose \textbf{Sample-Aligned Decoders (\method)}, a post-training framework that fine-tunes only the decoder of a frozen generator (\cref{fig:method}).
To resolve unpaired supervision, \emph{Generated-Sample Distribution Alignment} (GSA) decodes a one-time cache of generated latents and matches them to unpaired real images in a frozen DINOv2 feature space~\cite{oquab2023dinov2}.
To prevent latent-space drift, \emph{Encoder Anchor Preservation} (EAP) keeps the decoder close to the original one on encoder latents.
Our analysis explains why the alignment must act on generated latents. On encoder latents it is a zero-sum game against the anchor, whereas on generated latents it moves each sample towards real data in proportion to the density ratio between generated and encoder latents, so the decoder changes most where the generator deviates from the encoder distribution.
\method thus amortizes a discriminator-driven refinement of every sample into the decoder weights, learning the correction once, in about one GPU-hour, and applying it to every sample at no cost.
Because the anchor bounds how far each sample can move, the analysis also shows that \method corrects texture, color and local statistics but not missing diversity.

\begin{table}[t]
\centering
\footnotesize
\setlength{\tabcolsep}{3pt}
\newcommand{\cm}{\ding{51}}\newcommand{\xm}{\ding{55}}
\caption{\textbf{Comparison with related correction methods.} Generated latents: trained on latents produced by the generator ($\circ$: only partially generated, \ie teacher-forced or partially denoised). Reconstruction: the decoder stays valid on encoder latents.}
\label{tab:related}
\vspace{2pt}
\begin{tabular}{lcccc}
\toprule
\textbf{Method} & \makecell{Generated\\latents} & \makecell{No paired\\targets} & \makecell{Keeps\\reconstr.} & \makecell{No inference\\cost} \\
\midrule
Diffusion dec.~\cite{openai2023consistencydecoder,chen2025dito} & \xm & \xm & \cm & \xm \\
Noise-robust dec.~\cite{zheng2025rae} & \xm & \xm & \cm & \cm \\
RobusTok~\cite{qiu2025robustok} & $\circ$ & \xm & \cm & \cm \\
GAR~\cite{fang2026gar} & $\circ$ & \xm & \xm & \cm \\
FDDA~\cite{fang2026fdda} (contemp.) & \cm & \cm & \xm & \cm \\
Disc.\ guidance~\cite{kim2023dg} & \cm & \cm & -- & \xm \\
\rowcolor{oursbg} \method (ours) & \cm & \cm & \cm & \cm \\
\bottomrule
\end{tabular}
\end{table}

\method improves every generator we tested (\cref{fig:teaser}b), four state-of-the-art ImageNet generators across three tokenizers, without degrading reconstruction, and extends to text-to-image generation.
On REPA-E~\cite{leng2025repae}, one of the strongest public generators, it lowers FID-50K from $1.15$ to $0.98$.
Because \method only replaces the decoder, it is complementary to generator-side methods such as discriminator guidance~\cite{kim2023dg}, and it outperforms alternative ways of adapting the decoder.
In summary, our contributions are as follows:
\begin{itemize}
\item We show that generated latents lie on the encoder manifold but follow a different distribution, so the decoder faces a distribution shift and not a perturbation.
\item We propose \method, a plug-and-play post-training framework whose two modules, GSA and EAP, align the decoder of a frozen generator with its own samples at no inference cost, and show theoretically and empirically that the alignment must act on generated latents, where it becomes a density-ratio-weighted correction.
\item Experiments on four state-of-the-art ImageNet generators with three tokenizers and on text-to-image generation show consistent gains over the original decoders and over alternative decoder adaptations.
\end{itemize}

\section{Related Work}
\label{sec:related}

\noindent\textbf{Improving the latent space and the generator.}
Visual tokenizers compress images into discrete~\cite{esser2021taming,tian2024var,yu2024titok} or continuous~\cite{kingma2014vae,rombach2022ldm,li2024mar} latents, on which autoregressive~\cite{sun2024llamagen,tian2024var}, masked~\cite{li2024mar} or diffusion and flow models~\cite{peebles2023dit,ma2024sit} are trained.
Most progress improves this pipeline before or during generator training: by increasing the compression ratio~\cite{chen2025dcae}, aligning latents with vision foundation models~\cite{yao2025vavae,chen2025maetok,shi2025rectok}, making them easier to diffuse~\cite{skorokhodov2025diffusability}, aligning the generator's features~\cite{yu2025repa}, or tuning the autoencoder and the generator end to end~\cite{leng2025repae}.
Others update a trained generator by distillation~\cite{sauer2023add,yin2024dmd} or preference optimization~\cite{wallace2024dpo,bai2026tgo,lei2025psalign}, or reduce the exposure bias between its training and sampling inputs~\cite{bengio2015scheduled,ning2023input,ning2024elucidating,li2024alleviating}.
All of them train or update the generator, while the decoder is still trained only on encoder latents, so the mismatch between the generator and the decoder remains.
\method keeps the generator fixed and applies on top of these methods, as we show for VA-VAE and the E2E-VAE of REPA-E.

\rhead{Correcting samples at inference.}
Classifier-free and related guidance~\cite{ho2022cfg,kynkaanniemi2024interval,karras2024autoguidance} improves samples at sampling time.
Discriminator-based methods reject samples~\cite{azadi2019drs}, add discriminator gradients at every sampling step~\cite{kim2023dg}, or refine samples after sampling~\cite{tanaka2019dot,ansari2021dgflow}, all at extra test-time cost that grows with the number of samples and sampling steps.
\method follows the idea of amortizing an iterative refinement into a network~\cite{johnson2018cfg,yi2023monoflow}, learning the discriminator-driven correction once in the decoder weights, so inference cost is unchanged and the correction is complementary to sampling-time corrections such as discriminator guidance.

\rhead{Adapting the decoder.}
Closest to our work are methods that change the decoder itself.
A first family replaces or re-trains it for better reconstruction: the fine-tuned SD-VAE decoder~\cite{rombach2022ldm}, the Consistency Decoder~\cite{openai2023consistencydecoder,song2023consistency}, and diffusion-based decoders~\cite{zhao2024epsvae,chen2025dito}, which improve detail at the cost of multi-step decoding. All of them are trained on encoder latents and do not see the errors of a generator.
A second family makes the decoder robust to perturbed latents, assuming that generator errors resemble such perturbations: decoders are trained with noise added to the latents~\cite{zheng2025rae}, or post-trained with \emph{paired} targets from teacher-forced token sequences or SDEdit-perturbed real images~\cite{qiu2025robustok,meng2022sdedit} and from partially denoised latents~\cite{fang2026gar}, the latter motivated by a bound of generation FID by reconstruction FID and a latent Fréchet distance; RankE~\cite{jian2026ranke} co-trains a discrete decoder with the generator policy.
Contemporaneous with our work, FDDA~\cite{fang2026fdda} adapts the decoder on fully generated latents with a Fréchet objective over several feature spaces, including the Inception network used by FID, and without a term that preserves reconstruction.
\method differs in three respects: its alignment is \emph{unpaired} and acts on fully generated latents, it is balanced by an anchor that keeps the decoder valid on encoder latents, and it is supported by an analysis of where the alignment must act. \Cref{tab:related} summarizes these differences against the closest correction methods.

\rhead{Distribution matching in feature space.}
Matching distributions in pretrained feature spaces is effective for training generators~\cite{sauer2021projected,sauer2022styleganxl,li2017mmdgan,mathiasen2020fidbackprop}, but optimizing the space of an evaluation metric can improve the metric without improving images~\cite{kynkaanniemi2023role,stein2023exposing}; \method therefore aligns in DINOv2 space and is evaluated with Inception and CLIP features.

\section{Method}
\label{sec:method}

\input{fig/method}

As illustrated in \cref{fig:method}, \method fine-tunes only the decoder of a frozen latent generator and comprises two modules: (a)~\emph{Generated-Sample Distribution Alignment} (GSA), which aligns decoded generated latents with unpaired real images (\cref{sec:method:gsa}), and (b)~\emph{Encoder Anchor Preservation} (EAP), which keeps the decoder valid on the latent space it must decode (\cref{sec:method:eap}). We formalize the problem and then analyze why alignment must act on generated latents.

\subsection{Problem Formulation}
\label{sec:method:problem}
Let $E$ and $D_0$ be the encoder and decoder of a pretrained autoencoder, and let $p_E$ be the distribution of encoder latents $E(x)$, $x\sim p_{\mathrm{data}}$.
A generator $G$, trained on $p_E$ and run with a fixed sampler, produces latents $z\sim q$ that are decoded into images $D_0(z)$.
For a decoder $D_\phi$, let $p_{D_\phi}$ denote the distribution of decoded samples $D_\phi(z)$, $z\sim q$.
We compare image distributions in the feature space of a frozen network $\psi$ with an integral probability metric~\cite{muller1997ipm,sriperumbudur2012ipm},
\begin{equation}
d(\mu,\nu) = \sup_{f\in\mathcal F}\ \mathbb E_{x\sim\mu}\, f(\psi(x)) - \mathbb E_{x\sim\nu}\, f(\psi(x)),
\label{eq:ipm}
\end{equation}
which covers the maximum mean discrepancy (MMD)~\cite{gretton2012kernel} and, with a learned $f$, GAN discriminators.
Autoencoder training reduces the reconstruction gap and generator training brings $q$ close to $p_E$, but neither controls $d(p_{D_\phi}, p_{\mathrm{data}})$, the quantity that sample metrics measure. Once the generator is trained, the decoder is the last component that can still change it without retraining the generator or modifying the sampler. We therefore seek
\begin{equation}
\min_\phi\ d(p_{D_\phi},\, p_{\mathrm{data}})
\quad\text{s.t.}\quad
\mathbb E_{z\sim p_E}\|D_\phi(z) - D_0(z)\| \le \epsilon ,
\label{eq:constrained}
\end{equation}
where the objective is addressed by GSA and the trust-region constraint by EAP.

\subsection{Generated-Sample Distribution Alignment}
\label{sec:method:gsa}
No generated latent has a target image, but the population of decoded samples can be compared with the population of real images, so GSA resolves the lack of paired targets by aligning the two distributions.

\phead{Latent caching.} We sample $N$ generated latents $\{z_{\text{gen},i}\}$ once with the sampler of the deployed model ($16$k--$40$k, depending on the generator) and store them. The cache is an empirical sample of $q$, the distribution the decoder actually faces. Training then never runs or back-propagates through the generator, so its cost is independent of the generator size and the number of sampling steps.

\phead{Feature-space alignment.} The decoded latents $D_\phi(\zgen)$ and a set of unpaired real images are mapped by a frozen DINOv2-S~\cite{oquab2023dinov2} $\psi$, and GSA minimizes a differentiable estimate $\widehat d$ of \cref{eq:ipm}:
\begin{equation}
\Ldist(\phi) = \widehat d\,\big(\psi(D_\phi(\zgen)),\ \psi(x_{\mathrm{real}})\big).
\label{eq:gsa}
\end{equation}
We instantiate $\widehat d$ with three discrepancies that differ in how the witness function $f$ of \cref{eq:ipm} is obtained.
\emph{GAN} learns it: lightweight heads $h$ on four blocks of the frozen DINOv2~\cite{sauer2021projected,sauer2023add} are trained jointly with the decoder. For decoded and real features $\hat y=\psi(D_\phi(z))$ and $y_r=\psi(x_r)$, the heads minimize
\begin{equation}
\mathcal L_h = \mathbb E\big[1-h(y_r)\big]_+ + \mathbb E\big[1+h(\hat y)\big]_+,
\label{eq:gan}
\end{equation}
and the decoder minimizes $\Ldist = -\mathbb E\, h(\hat y)$.
\emph{MMD} fixes it in a kernel space: $\Ldist$ is the unbiased estimate of $\mathrm{MMD}^2$ with a sum of Gaussian kernels~\cite{gretton2012kernel,li2017mmdgan}, computed between a batch of decoded features and precomputed real features.
\emph{FD} matches the first two moments~\cite{mathiasen2020fidbackprop}, with the generated statistics estimated from a queue of recent features:
\begin{equation}
\Ldist = \|\mu_g-\mu_r\|^2 + \mathrm{Tr}\big(\Sigma_g+\Sigma_r-2(\Sigma_g\Sigma_r)^{1/2}\big).
\label{eq:fd}
\end{equation}
We align in DINOv2 space because its patch features capture both semantics and local texture, where the residual errors of the generator appear, and differ from the evaluation spaces~\cite{kynkaanniemi2023role}.

\subsection{Encoder Anchor Preservation}
\label{sec:method:eap}
Without paired targets, GSA alone lets the decoder drift, since it can satisfy the feature-space objective by forgetting how to decode the latent space. EAP replaces the constraint in \cref{eq:constrained} with a soft anchor on encoder latents,
\begin{equation}
\Lanc(\phi) = \mathbb{E}_{x\sim p_{\mathrm{data}}} \big\| D_\phi(E(x)) - D_0(E(x)) \big\|_1 ,
\label{eq:eap}
\end{equation}
which preserves reconstruction and keeps the decoder valid on the encoder latent space on which the generator was trained.
Two design choices matter. First, the anchor targets the output of $D_0$ and not $x$. It thus requires the decoder to keep its behavior, not to reconstruct better, and does not push the decoder along the trade-off between pixel fidelity and realism that reconstruction losses induce. Second, it acts on encoder latents of unlabeled real images, while GSA acts on generated latents. The two losses therefore constrain the decoder on different parts of the latent space, and the decoder remains free to change where generated latents are more likely than encoder latents (\cref{sec:method:analysis}).

\phead{Optimization.} The final objective of \method is
\begin{equation}
\mathcal{L}(\phi) = \Lanc(\phi) + \lambda\, \Ldist(\phi).
\label{eq:sad}
\end{equation}
The decoder $D_\phi$ is initialized at $D_0$, and each step decodes a batch of cached latents for $\Ldist$ and a batch of encoder latents for $\Lanc$. We update only the decoder for 2{,}000 steps, which takes about one GPU-hour, and at inference $D_\phi$ simply replaces $D_0$, so the sampling pipeline and its cost are unchanged (\cref{alg:sad} in the appendix).

\begin{table*}[t]
\centering
\footnotesize
\setlength{\tabcolsep}{6pt}
\renewcommand{\arraystretch}{1.02}
\caption{\textbf{Comparison with state-of-the-art generators on ImageNet $256{\times}256$.} With guidance; FID-50K, ADM evaluator. For each released generator, the first row uses its original decoder $D_0$ and the second row \method (GAN discrepancy) on identical latents. Bold marks a result no worse than $D_0$. \fdclip{} uses a held-out reference and is computed by us, for other generators from their released samples or weights. $^\dagger$Class-balanced labels.}
\label{tab:sota}
\vspace{2pt}
\begin{tabular}{llcccccc}
\toprule
\textbf{Method} & \textbf{Tokenizer} & FID$\downarrow$ & sFID$\downarrow$ & IS$\uparrow$ & Prec.$\uparrow$ & Rec.$\uparrow$ & \fdclip$\downarrow$ \\
\midrule
\multicolumn{8}{l}{\textit{Other generators}} \\
ADM-G~\cite{dhariwal2021adm} & -- & 4.59 & 5.25 & 186.7 & 0.82 & 0.52 & 46.9 \\
VAR-d30~\cite{tian2024var} & VQVAE & 1.92 & -- & 323.1 & 0.82 & 0.59 & 54.0 \\
MAR-H~\cite{li2024mar} & KL-16 & 1.55 & -- & 303.7 & 0.81 & 0.62 & -- \\
SiT-XL/2~\cite{ma2024sit} & SD-VAE & 2.06 & 4.50 & 270.3 & 0.82 & 0.59 & 44.3 \\
MDTv2-XL/2~\cite{gao2023mdtv2} & SD-VAE & 1.58 & 4.52 & 314.7 & 0.79 & 0.65 & -- \\
\midrule
DiT-XL/2~\cite{peebles2023dit} & SD-VAE & 2.30 & 4.51 & 277.8 & 0.83 & 0.57 & 42.8 \\
\rowcolor{oursbg} \quad + \method (ours) & & \textbf{2.13}\,{\scriptsize\textcolor{gaincol}{$-$0.17}} & \textbf{4.25}\,{\scriptsize\textcolor{gaincol}{$-$0.26}} & \textbf{280.5}\,{\scriptsize\textcolor{gaincol}{+2.7}} & \textbf{0.83}\,{\scriptsize\textcolor{gaincol}{+0.00}} & \textbf{0.58}\,{\scriptsize\textcolor{gaincol}{+0.01}} & \textbf{36.8}\,{\scriptsize\textcolor{gaincol}{$-$6.0}} \\
\midrule
DiT-XL/2 + DG~\cite{kim2023dg} & SD-VAE & 1.83 & 5.13 & 285.3 & 0.78 & 0.63 & 46.4 \\
\rowcolor{oursbg} \quad + \method (ours) & & \textbf{1.69}\,{\scriptsize\textcolor{gaincol}{$-$0.14}} & \textbf{4.24}\,{\scriptsize\textcolor{gaincol}{$-$0.89}} & \textbf{286.2}\,{\scriptsize\textcolor{gaincol}{+0.9}} & \textbf{0.79}\,{\scriptsize\textcolor{gaincol}{+0.01}} & \textbf{0.63}\,{\scriptsize\textcolor{gaincol}{+0.00}} & \textbf{41.0}\,{\scriptsize\textcolor{gaincol}{$-$5.4}} \\
\midrule
SiT-XL/2 + REPA~\cite{yu2025repa} & SD-VAE & 1.43 & 4.67 & 306.7 & 0.79 & 0.64 & 39.9 \\
\rowcolor{oursbg} \quad + \method (ours) & & \textbf{1.38}\,{\scriptsize\textcolor{gaincol}{$-$0.05}} & \textbf{4.00}\,{\scriptsize\textcolor{gaincol}{$-$0.67}} & 305.4\,{\scriptsize\textcolor{losscol}{$-$1.3}} & \textbf{0.79}\,{\scriptsize\textcolor{gaincol}{+0.00}} & \textbf{0.64}\,{\scriptsize\textcolor{gaincol}{+0.00}} & \textbf{39.4}\,{\scriptsize\textcolor{gaincol}{$-$0.5}} \\
\midrule
LightningDiT-XL~\cite{yao2025vavae} & VA-VAE & 1.38 & 4.18 & 291.4 & 0.79 & 0.65 & 29.4 \\
\rowcolor{oursbg} \quad + \method (ours) & & \textbf{1.23}\,{\scriptsize\textcolor{gaincol}{$-$0.15}} & \textbf{3.82}\,{\scriptsize\textcolor{gaincol}{$-$0.36}} & \textbf{294.4}\,{\scriptsize\textcolor{gaincol}{+3.0}} & \textbf{0.80}\,{\scriptsize\textcolor{gaincol}{+0.01}} & \textbf{0.65}\,{\scriptsize\textcolor{gaincol}{+0.00}} & \textbf{28.2}\,{\scriptsize\textcolor{gaincol}{$-$1.2}} \\
\midrule
REPA-E~\cite{leng2025repae}$^\dagger$ & E2E-VAE & 1.15 & 4.10 & 301.5 & 0.79 & 0.67 & 29.1 \\
\rowcolor{oursbg} \quad + \method (ours) & & \textbf{0.98}\,{\scriptsize\textcolor{gaincol}{$-$0.17}} & \textbf{3.74}\,{\scriptsize\textcolor{gaincol}{$-$0.36}} & \textbf{304.4}\,{\scriptsize\textcolor{gaincol}{+2.9}} & \textbf{0.79}\,{\scriptsize\textcolor{gaincol}{+0.00}} & \textbf{0.67}\,{\scriptsize\textcolor{gaincol}{+0.00}} & \textbf{26.4}\,{\scriptsize\textcolor{gaincol}{$-$2.7}} \\
 
\bottomrule
\end{tabular}
\end{table*}

\subsection{Why Align on Generated Latents}
\label{sec:method:analysis}
The two losses act on different latents: EAP on $p_E$ and GSA on $q$. We analyze the consequence to first order, treating the decoder as $D_\phi = D_0 + \delta$ with an $\ell_2$ anchor and the MMD in the feature space of $\psi$, whose witness function $w$ has a gradient $\nabla w$ that points towards real data~\cite{gretton2012kernel,arbel2019mmdflow}. Proofs are in \cref{sec:proofs}.

\needspace{6\baselineskip}
\begin{proposition}[Encoder latents: a zero-sum game]
\label{prop:enc}
If the alignment is applied to encoder latents ($q = p_E$), every decoder that satisfies the anchor exactly has the same alignment loss as $D_0$, \ie $\Lanc(\phi)=0 \Rightarrow \Ldist(\phi) = \Ldist(\phi_0)$.
\end{proposition}

\needspace{6\baselineskip}
\begin{proposition}[Density-ratio-weighted correction]
\label{prop:opt}
Let $r(z) = q(z)/p_E(z)$. To first order in $\delta$, the minimizer of $\mathbb E_{p_E}\|\delta\|^2 + \lambda\,\mathrm{MMD}_\psi^2(p_{D_0+\delta},\, p_{\mathrm{data}})$ is
\begin{equation}
\delta^\star(z) = -\lambda\, r(z)\, J_\psi^{\top}\, \nabla w\big(\psi(D_0(z))\big),
\label{eq:optimal}
\end{equation}
where $J_\psi$ is the Jacobian of $\psi$ at $D_0(z)$.
\end{proposition}

\noindent \Cref{prop:enc} explains why aligning encoder latents can only trade reconstruction for nothing. \Cref{prop:opt} shows that on generated latents each decoded sample moves towards real data in proportion to $r(z)$, so the decoder changes most where the generator places more mass than the encoder, which is where its errors are. \method thus amortizes discriminator-driven sample refinement~\cite{tanaka2019dot,ansari2021dgflow} into the decoder weights at no inference cost. Because the anchor bounds how far a sample can move, \method repairs local errors such as texture but cannot restore missing diversity (see \cref{sec:proofs} for the proof).

\phead{Relation to discriminator guidance.} Discriminator guidance~\cite{kim2023dg} estimates the log density ratio between real and generated data with a discriminator and adds its gradient to the score at every sampling step, so it corrects the latent trajectory at inference time. \Cref{prop:opt} shows that \method applies the same kind of discriminator-driven correction, but to the decoded image and amortized into the decoder weights. The two act on different stages of the pipeline, the latent and its decoding, and can therefore be combined.

\section{Experiments}
\label{sec:exp}

We compare \method with state-of-the-art generators and alternative corrections, including text-to-image generation (\cref{sec:exp:comparison}), ablate its modules (\cref{sec:exp:ablation}), show qualitative results (\cref{sec:exp:case}) and analyze it (\cref{sec:exp:analysis}).

\subsection{Experimental Setup}
\label{sec:exp:setup}

\noindent\textbf{Generators.} We apply \method to four class-conditional ImageNet $256{\times}256$ generators with three tokenizers: DiT-XL/2~\cite{peebles2023dit} and SiT-XL/2 with REPA~\cite{yu2025repa} on SD-VAE, LightningDiT-XL~\cite{yao2025vavae} on VA-VAE, and REPA-E~\cite{leng2025repae} on its end-to-end tuned E2E-VAE, each with its released checkpoint and official sampling recipe; we also combine DiT-XL/2 with discriminator guidance (DG)~\cite{kim2023dg}. For text-to-image generation we use PixArt-$\alpha$~\cite{chen2024pixart} on COCO.

\phead{Evaluation settings and metrics.} We evaluate in two settings. The \emph{official setting} uses each released checkpoint with its official sampling recipe and reports FID-50K, sFID, Inception Score, precision and recall with the ADM evaluator~\cite{dhariwal2021adm}. The \emph{ablation setting} uses DiT-XL/2 with 50-step DPM-Solver++~\cite{lu2022dpmsolverpp}, decodes 10k latents and evaluates against a held-out half of the ADM reference batch. Since GSA is trained in DINOv2 space, we additionally report the Fréchet distance in CLIP space (\fdclip), which plays no role in training~\cite{kynkaanniemi2023role,stein2023exposing}, and measure reconstruction by rFID and PSNR on 10k ImageNet images.

\phead{Implementation details.} The cached training latents (16k--40k per generator) use seeds disjoint from evaluation. GSA uses 5k real images from the other half of the ADM batch, and EAP uses ImageNet training images. We train for 2{,}000 steps with AdamW at a learning rate of $10^{-5}$, with $\lambda{=}0.05$ for GAN and $\lambda{=}1$ for MMD, on one NVIDIA L20. These settings are shared across ImageNet generators (GAN by default); text-to-image uses a smaller $\lambda$.

\subsection{Performance Comparison}
\label{sec:exp:comparison}

\noindent\textbf{Comparison with state-of-the-art generators.} We first ask whether \method still helps the strongest generators, whose residual errors are already small. \Cref{tab:sota} shows that it does:
\textbf{1)} \method improves every generator and every tokenizer. It lowers FID-50K from $1.15$ to $0.98$ on REPA-E, below $1$, from $1.38$ to $1.23$ on LightningDiT and from $1.43$ to $1.38$ on SiT with REPA.
\textbf{2)} The gain stacks on top of discriminator guidance (DG), since decoding the samples of its released model with \method, trained only on plain DiT samples, further lowers FID-50K from $1.83$ to $1.69$ and sFID from $5.13$ to $4.24$ at no extra cost.
\textbf{3)} The gain is not a trade against other aspects of the samples, as the held-out \fdclip{} and sFID improve together with FID for every generator, precision and recall are maintained, and rFID changes by at most $0.13$. KID and the SigLIP Fréchet distance, also unused in training, agree (\cref{tab:extra_metrics} in the appendix).
The decoder thus corrects residual errors that remain even in the strongest generators instead of fitting the evaluation metric.

\begin{table}[t]
\centering
\footnotesize
\setlength{\tabcolsep}{3.6pt}
\caption{\textbf{Comparison of generator-side and decoder-side correction.} DiT-XL/2 in the official setting at guidance 1.5 (FID-50K, ADM evaluator). The best results are in bold and the second best are underlined.}
\label{tab:dg}
\vspace{2pt}
\begin{tabular}{lccccc}
\toprule
\textbf{Method} & \makecell{Extra\\sampling} & \makecell{Disc. real\\images} & FID$\downarrow$ & sFID$\downarrow$ & IS$\uparrow$ \\
\midrule
$D_0$ & -- & -- & 2.30 & 4.51 & \underline{277.8} \\
DG~\cite{kim2023dg} & $+$20\% & 1.28M & \underline{2.01} & 4.49 & 266.5 \\
\rowcolor{oursbg} \method (ours) & 0 & 5k & 2.13 & \underline{4.25} & \textbf{280.5} \\
\rowcolor{oursbg} DG + \method (ours) & $+$20\% & 1.28M & \textbf{1.86} & \textbf{4.02} & 269.3 \\
 
\bottomrule
\end{tabular}
\end{table}

\phead{Comparison with discriminator guidance.} DG is the closest generator-side counterpart of \method: both correct samples with a discriminator between generated and real data, but DG adds its gradient to the latent at every sampling step, whereas \method learns the correction once in the decoder. \Cref{tab:dg} compares them at the same guidance scale. \textbf{1)} \method obtains more than half of the FID gain of DG ($2.30\!\to\!2.13$ against $2.30\!\to\!2.01$) without extra sampling cost and with 5k real images instead of the 1.28M ImageNet training images used by the DG discriminator. \textbf{2)} \method improves sFID and IS together with FID, whereas DG lowers IS by 11 points. \textbf{3)} Combined, they reach FID $1.86$ and sFID $4.02$, the best at this scale. The two corrections act on the latent and on its decoding, respectively, and can therefore be combined.

\begin{table}[t]
\centering
\footnotesize
\setlength{\tabcolsep}{4.5pt}
\caption{\textbf{Comparison with decoder adaptation methods.} DiT-XL/2 in the ablation setting. Noise-robust, paired and FDDA are our re-implementations. The best results are in bold and the second best are underlined. $^\ast$Trained with Inception, the FID network.}
\label{tab:adapt}
\vspace{2pt}
\begin{tabular}{lcccc}
\toprule
\textbf{Method} & FID$\downarrow$ & \fdclip$\downarrow$ & rFID$\downarrow$ & PSNR$\uparrow$ \\
\midrule
Original decoder $D_0$ & 13.32 & 51.4 & 2.63 & 23.91 \\
\midrule
\multicolumn{5}{l}{\textit{Replacing or re-training the decoder}} \\
sd-vae-ft-mse~\cite{rombach2022ldm} & 13.40 & 51.8 & 2.91 & \textbf{24.75} \\
Consistency Decoder~\cite{openai2023consistencydecoder} & 13.18 & 57.3 & 3.34 & 23.31 \\
Fine-tuning, $\ell_1$+LPIPS+GAN & 13.40 & 48.6 & 7.02 & \underline{24.51} \\
\midrule
\multicolumn{5}{l}{\textit{Robustness to perturbed latents}} \\
Noise-robust decoder~\cite{zheng2025rae} & 13.40 & 51.6 & 2.67 & 24.02 \\
Paired post-training~\cite{qiu2025robustok,fang2026gar} & 13.10 & 50.8 & 2.59 & 23.91 \\
\midrule
\multicolumn{5}{l}{\textit{Alignment to real images}} \\
Pixel refiner after $D_0$ & 12.54 & 45.3 & 2.63 & 23.86 \\
FDDA~\cite{fang2026fdda}$^\ast$ & \textbf{12.21} & 52.1 & 5.91 & 21.16 \\
\midrule
\rowcolor{oursbg} \method-GAN (ours) & \underline{12.36} & \underline{42.4} & \textbf{2.56} & 23.92 \\
\rowcolor{oursbg} \method-MMD (ours) & 12.68 & \textbf{41.6} & \underline{2.57} & 23.92 \\
 
\bottomrule
\end{tabular}
\end{table}

\phead{Comparison with decoder adaptation methods.} To test whether the gain requires the design of \method or whether any change to the decoder would do, \cref{tab:adapt} compares decoders on the same latents.
\textbf{1)} Better reconstruction does not give better samples, as public decoders for the same latent space~\cite{rombach2022ldm,openai2023consistencydecoder} and fine-tuning toward $x$ leave FID at $13.18$--$13.40$.
\textbf{2)} Robustness to perturbed latents does not help either. Noise-robust decoders~\cite{zheng2025rae} and paired post-training~\cite{qiu2025robustok} are trained on random or SDEdit perturbations of real latents, which do not match the distribution of generated latents, and stay within $0.22$ of $D_0$.
\textbf{3)} Aligning with real images is what matters, and \method does it best. A pixel refiner trained with the same GSA objective is weaker and adds 15\% to the decoding cost, and FDDA~\cite{fang2026fdda}, which optimizes the Inception features of FID directly, lowers FID but worsens \fdclip{} and more than doubles rFID. \method achieves the best \fdclip{} and rFID, and the same ranking holds in the official setting (\cref{tab:adapt_official}).

\begin{table}[t]
\centering
\footnotesize
\setlength{\tabcolsep}{3pt}
\caption{\textbf{Text-to-image generation.} PixArt-$\alpha$ on COCO (10k samples). Pick: share of samples preferred by PickScore over $D_0$ (same latent; for real val2017 images, same caption). All tested weights are in \cref{sec:supp:t2i}.}
\label{tab:t2i}
\vspace{2pt}
\begin{tabular}{lcccccc}
\toprule
\textbf{Decoder} & FID$\downarrow$ & \fdclip$\downarrow$ & CLIP$\uparrow$ & Rec.$\uparrow$ & rFID$\downarrow$ & Pick \\
\midrule
$D_0$ & 34.73 & 209.1 & 26.03 & 0.286 & 4.28 & -- \\
\rowcolor{oursbg} GAN, $\lambda{=}0.01$ & 23.55 & 141.6 & 26.91 & 0.416 & 4.29 & 48\% \\
\rowcolor{oursbg} MMD, $\lambda{=}0.1$ & 28.35 & 150.4 & 26.19 & 0.343 & 4.26 & 67\% \\
\midrule
Real images & -- & -- & 25.69 & -- & -- & 18\% \\
 
\bottomrule
\end{tabular}
\end{table}

\phead{Text-to-image generation.} We next ask whether \method extends beyond class-conditional generation. \Cref{tab:t2i} applies it to PixArt-$\alpha$~\cite{chen2024pixart} at $256{\times}256$, with COCO training captions for the cached latents and COCO images as the GSA target. \textbf{1)} \method lowers FID by 18--32\% and \fdclip{} by 28--32\%, and raises recall from $0.286$ to $0.343$--$0.416$, at unchanged rFID. \textbf{2)} Text alignment is preserved, with the CLIP score rising from $26.03$ to $26.19$--$26.91$. \textbf{3)} PickScore prefers \method in 48--67\% of the samples. A larger $\lambda$ lowers FID but also this preference (\cref{sec:supp:t2i}), although PickScore does not measure realism, as real images win only 18\% against $D_0$.

\subsection{Ablation Study}
\label{sec:exp:ablation}

\begin{table}[t]
\centering
\footnotesize
\setlength{\tabcolsep}{2.6pt}
\caption{\textbf{Ablation study on the modules of \method.} DiT-XL/2 in the ablation setting.}
\label{tab:abl_main}
\vspace{2pt}
\begin{tabular}{clcccc}
\toprule
\textbf{\#} & \textbf{Variant} & FID$\downarrow$ & \fdclip$\downarrow$ & rFID$\downarrow$ & PSNR$\uparrow$ \\
\midrule
0 & $D_0$ & 13.32 & 51.4 & 2.63 & 23.91 \\
\midrule
\multicolumn{6}{l}{\textit{Ablation on GSA}} \\
1 & w/o GSA (EAP only) & 13.31 & 49.9 & 2.65 & \textbf{23.92} \\
2 & GSA on $E(x)$ instead of $\zgen$ & 13.59 & 51.5 & 7.70 & 23.90 \\
3 & GSA with MMD & 12.68 & \textbf{41.6} & 2.57 & 23.92 \\
4 & GSA with FD & 12.79 & 41.8 & 2.59 & 23.91 \\
\midrule
\multicolumn{6}{l}{\textit{Ablation on EAP}} \\
5 & w/o EAP & 13.06 & 50.3 & 5.06 & 21.92 \\
6 & w/o EAP, lower lr ($3{\times}10^{-6}$) & 12.52 & 50.2 & 4.65 & 21.94 \\
7 & w/o EAP, early stop (500 steps) & 12.63 & 55.6 & 5.12 & 22.48 \\
\midrule
\rowcolor{oursbg} & \method (GSA + EAP) & \textbf{12.36} & 42.4 & \textbf{2.56} & 23.92 \\
 
\bottomrule
\end{tabular}
\end{table}

\Cref{tab:abl_main} examines which parts of \method are necessary by changing one module at a time.
\textbf{Ablation on GSA.} Removing GSA (\#1) leaves FID at $13.31$, so GSA is the source of the gain. Where it is applied is decisive. On encoder latents (\#2) it worsens FID to $13.59$ and nearly triples rFID, as predicted by \cref{prop:enc}, and this placement fails for all three discrepancies and on all three official generators (FID-50K $2.30\!\to\!2.92$ on DiT-XL/2, $1.38\!\to\!2.28$ on LightningDiT and $1.15\!\to\!1.80$ on REPA-E). On generated latents all three discrepancies work (\#3, \#4), with GAN giving the lowest FID and MMD and FD the lowest \fdclip. The gain therefore comes from where the alignment acts rather than from a particular discrepancy.
\textbf{Ablation on EAP.} Without EAP (\#5), \fdclip{} returns to the level of $D_0$, rFID doubles and PSNR drops by $2$\,dB, and neither a lower learning rate (\#6) nor early stopping (\#7) recovers them; in the official setting, FID-50K rises from $2.13$ to $3.08$. EAP therefore not only preserves reconstruction but also keeps the decoder from overfitting the feature space of GSA.

\begin{figure}[t]
\centering
\begin{tikzpicture}
\pgfplotsset{sen/.style={sadplot, scale only axis, width=0.255\linewidth, height=0.25\linewidth, grid=none, ymajorgrids,
  tick label style={font=\footnotesize}, xlabel style={yshift=3pt, font=\footnotesize}, ylabel style={font=\footnotesize}},
  senR/.style={sen, axis y line*=right, axis x line=none, ymajorgrids=false, ytick pos=right, yticklabel pos=right, ylabel near ticks}}
\begin{axis}[sen, name=A, xmin=-0.4, xmax=4.4, xtick={0,1,2,3,4}, xticklabels={0,0.1,0.3,1,3},
  ymin=12.2, ymax=13.2, ytick={12.4,12.8,13.2}, axis y line*=left, ylabel={FID $\downarrow$}, ylabel style={text=cSAD, yshift=-6pt},
  xlabel={(a) EAP weight}]
  \addplot[cSAD, mark=*, mark size=1.6pt] table[x expr=\coordindex, y=fid] {fig/data/abl_anchor.dat};\label{se:fid}
\end{axis}
\begin{axis}[senR, xmin=-0.4, xmax=4.4, ymin=2, ymax=5.5, ytick={2,3,4,5}, ylabel={rFID $\downarrow$}, ylabel style={text=cDz, yshift=6pt}]
  \addplot[cDz, densely dashed, mark=square*, mark size=1.4pt] table[x expr=\coordindex, y=rfid] {fig/data/abl_anchor.dat};\label{se:rfid}
\end{axis}
\begin{axis}[sen, name=B, at={(A.south east)}, anchor=south west, xshift=0.245\linewidth, xmin=-0.4, xmax=3.4, xtick={0,1,2,3},
  xticklabels={0,0.05,0.1,0.2}, ymin=12.2, ymax=13.5, ytick={12.4,12.8,13.2}, axis y line*=left,
  ylabel={FID $\downarrow$}, ylabel style={text=cSAD, yshift=-6pt}, xlabel={(b) GSA weight $\lambda$}]
  \addplot[cSAD, mark=*, mark size=1.6pt] coordinates {(0,13.311) (1,12.363) (2,12.392) (3,12.344)};
\end{axis}
\begin{axis}[senR, at={(A.south east)}, anchor=south west, xshift=0.245\linewidth, xmin=-0.4, xmax=3.4, ymin=38, ymax=52,
  ytick={40,45,50}, ylabel={\fdclip $\downarrow$}, ylabel style={text=cRef, yshift=6pt}]
  \addplot[cRef, densely dashed, mark=triangle*, mark size=1.8pt] coordinates {(0,49.92) (1,42.43) (2,42.96) (3,45.90)};\label{se:clip}
\end{axis}
\node[font=\footnotesize, anchor=south] at ([yshift=1pt]current bounding box.north) {%
  \ref{se:fid}\,FID\qquad \ref{se:rfid}\,rFID\qquad \ref{se:clip}\,\fdclip};
\end{tikzpicture}
\vspace{-6pt}
\caption{\textbf{Sensitivity to the loss weights.} DiT-XL/2 in the ablation setting. (a) Weight of $\Lanc$ with $\lambda=0.05$. (b) Weight $\lambda$ of $\Ldist$ with the weight of $\Lanc$ fixed to 1. Defaults: 1 and 0.05.}
\label{fig:sensitivity}
\end{figure}
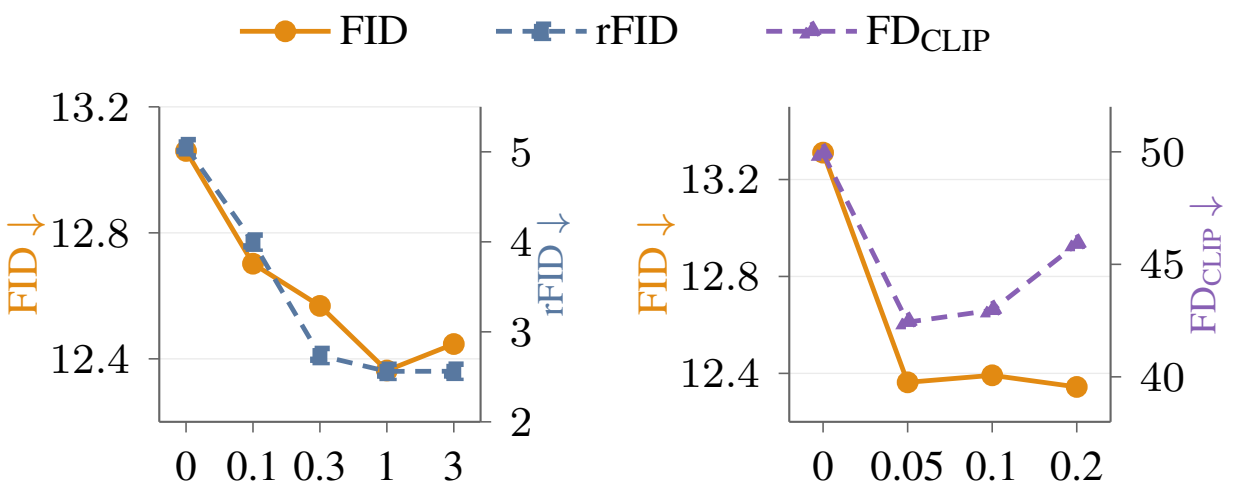

\phead{Sensitivity analysis.} \Cref{fig:sensitivity} shows how sensitive \method is to the two loss weights. As the EAP weight decreases from $1$ to $0$, FID and rFID degrade together, while a larger weight of $3$ changes little. FID is insensitive to the GSA weight $\lambda$ ($12.34$--$12.39$ over a $4\times$ range), whereas \fdclip{} prefers the default. \method is also fast and data-efficient, as 500 training steps, 1k cached latents or 500 real images already give most of the gain (\cref{fig:ablation} in the appendix).

\begin{figure*}[t]
\centering
\setlength{\tabcolsep}{0.5pt}
\renewcommand{\arraystretch}{0}
\def\qw{2.11cm}
\begin{tabular}{@{}c@{\hspace{2pt}}cccccccc@{}}
\rotatebox{90}{\makebox[\qw][c]{\scriptsize $D_0$}} & \includegraphics[width=\qw]{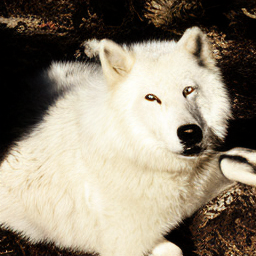} & \includegraphics[width=\qw]{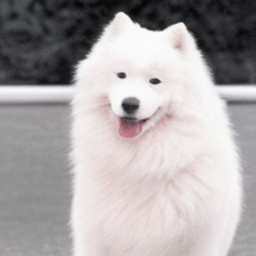} & \includegraphics[width=\qw]{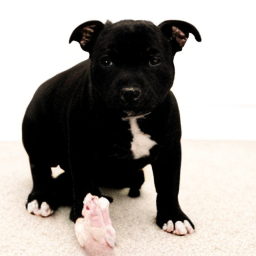} & \includegraphics[width=\qw]{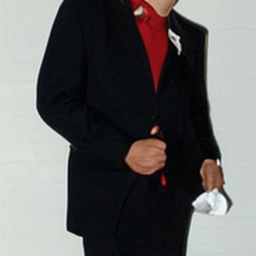} & \includegraphics[width=\qw]{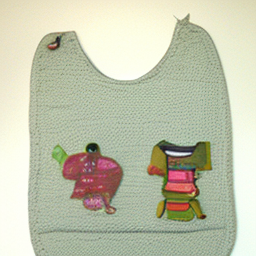} & \includegraphics[width=\qw]{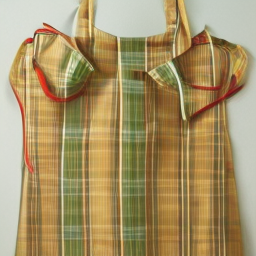} & \includegraphics[width=\qw]{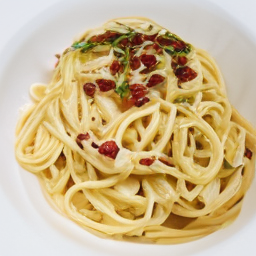} & \includegraphics[width=\qw]{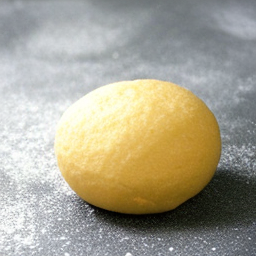} \\[1pt]
\rotatebox{90}{\makebox[\qw][c]{\scriptsize \textbf{\method}}} & \includegraphics[width=\qw]{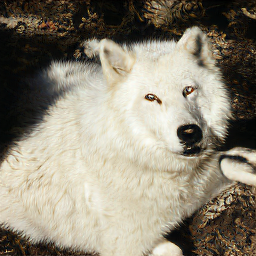} & \includegraphics[width=\qw]{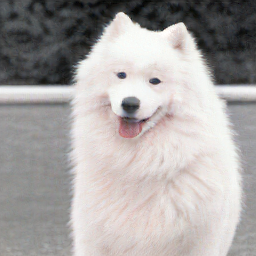} & \includegraphics[width=\qw]{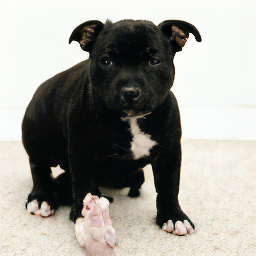} & \includegraphics[width=\qw]{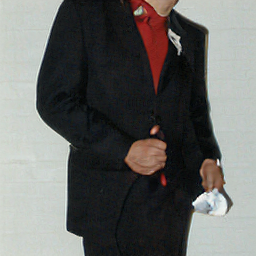} & \includegraphics[width=\qw]{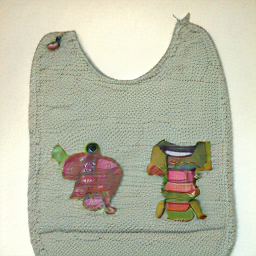} & \includegraphics[width=\qw]{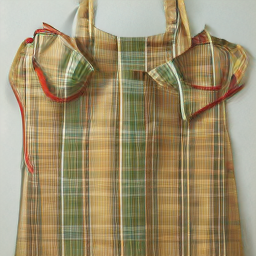} & \includegraphics[width=\qw]{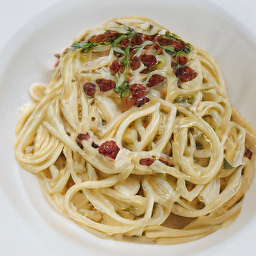} & \includegraphics[width=\qw]{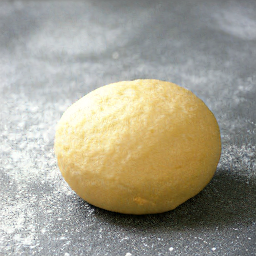} \\[1pt]
\rotatebox{90}{\makebox[\qw][c]{\scriptsize $|\text{\method}-D_0|$}} & \includegraphics[width=\qw]{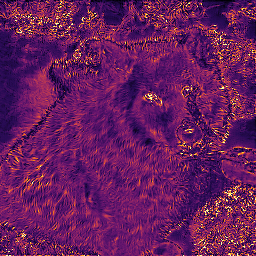} & \includegraphics[width=\qw]{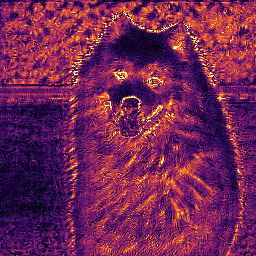} & \includegraphics[width=\qw]{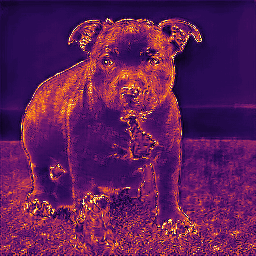} & \includegraphics[width=\qw]{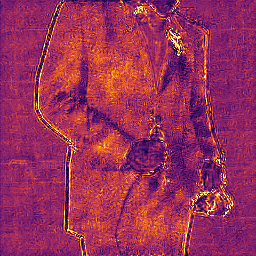} & \includegraphics[width=\qw]{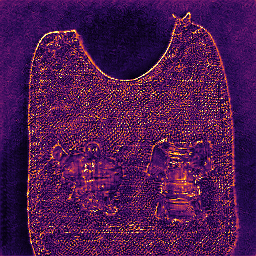} & \includegraphics[width=\qw]{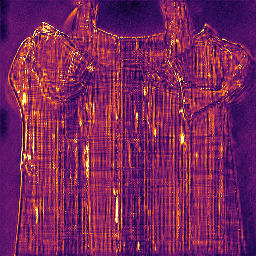} & \includegraphics[width=\qw]{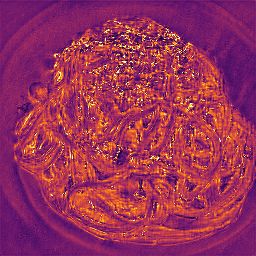} & \includegraphics[width=\qw]{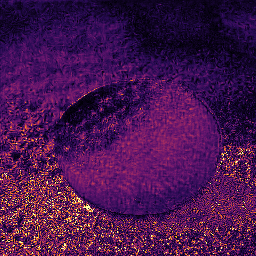} \\
\end{tabular}
\vspace{-4pt}
\caption{\textbf{Qualitative comparison on DiT-XL/2.} The same generated latents are decoded by $D_0$ and by \method. The bottom row shows their per-pixel absolute difference (brighter is larger).}
\label{fig:qual}
\end{figure*}

\subsection{Case Study}
\label{sec:exp:case}

\Cref{fig:qual} compares the same generated latents decoded by $D_0$ and by \method. Since the latent and the generator are unchanged, \method keeps the content and layout of each sample but changes how it is rendered. It reduces the color cast and overexposure of $D_0$, as on the white fur of the wolf and the Samoyed, and restores detail in saturated regions such as black fur and the dark suit. The difference maps show that the changes follow the objects and textured areas such as fabric, pasta and flour, and stay small on plain backgrounds. More results are in \cref{sec:supp:qual}.

\subsection{Analysis}
\label{sec:exp:analysis}

\begin{figure}[t]
\centering
\begin{tikzpicture}
\pgfplotsset{anm/.style={sadplot, scale only axis, width=0.335\linewidth, height=0.30\linewidth, tick label style={font=\footnotesize},
  xlabel style={yshift=3pt, font=\footnotesize}, ylabel style={yshift=-5pt, font=\footnotesize}}}
\begin{axis}[anm, name=A, clip mode=individual, xmode=log, xmin=2.45, xmax=9, ymin=12.15, ymax=14.15,
  xtick={2.5,4,7}, xticklabels={2.5,4,7}, log ticks with fixed point, minor xtick={}, ytick={12.5,13,13.5,14},
  xlabel={(a) rFID $\downarrow$ (log)}, ylabel={FID $\downarrow$}]
  \fill[cGood] (axis cs:2.45,12.15) rectangle (axis cs:2.9,13.319);
  \draw[black!45, densely dashed, line width=0.5pt] (axis cs:2.631,12.15) -- (axis cs:2.631,14.15);
  \draw[black!45, densely dashed, line width=0.5pt] (axis cs:2.45,13.319) -- (axis cs:9,13.319);
  \addplot[only marks, ptS] table[x=rfid, y=fid, discard if not={grp}{gen}] {fig/data/scatter.dat};\label{an:gen}
  \addplot[only marks, ptE] table[x=rfid, y=fid, discard if not={grp}{enc}] {fig/data/scatter.dat};\label{an:enc}
  \addplot[only marks, cNoise, mark=square*, mark size=1.4pt] table[x=rfid, y=fid, discard if not={grp}{noise}] {fig/data/scatter.dat};\label{an:oth}
  \addplot[only marks, cNoise, mark=square*, mark size=1.4pt] table[x=rfid, y=fid, discard if not={grp}{paired}] {fig/data/scatter.dat};
  \addplot[only marks, cNoise, mark=square*, mark size=1.4pt] table[x=rfid, y=fid, discard if not={grp}{refiner}] {fig/data/scatter.dat};
  \addplot[only marks, black, mark=star, mark size=3pt, line width=0.8pt] coordinates {(2.631,13.319)};\label{an:d0}
\end{axis}
\begin{axis}[anm, name=B, at={(A.south east)}, anchor=south west, xshift=0.17\linewidth, xmin=0, xmax=1, ymin=-0.4, ymax=0.12,
  xtick={0,0.5,1}, xticklabels={0,0.5,1}, ytick={-0.4,-0.2,0}, yticklabels={$-$0.4,$-$0.2,0},
  xlabel={(b) Spatial frequency}, ylabel={$\log_{10}(P/P_{\mathrm{real}})$}]
  \draw[black!50, line width=0.5pt] (axis cs:0,0) -- (axis cs:1,0);
  \addplot[black, densely dashed, no marks] table[x=f, y=r_orig] {fig/data/spectrum.dat};\label{an:l0}
  \addplot[cEnc, densely dotted, no marks, line width=1pt] table[x=f, y=r_enc] {fig/data/spectrum.dat};\label{an:le}
  \addplot[cSAD, no marks, line width=1.1pt] table[x=f, y=r_sad] {fig/data/spectrum.dat};\label{an:ls}
\end{axis}
\node[font=\footnotesize, anchor=south, align=center] at ([yshift=1pt]current bounding box.north) {%
  (a)\quad \ref{an:d0}\,$D_0$\quad \ref{an:gen}\,GSA on $\zgen$ (\method)\quad \ref{an:enc}\,GSA on $E(x)$\quad \ref{an:oth}\,others\\
  (b)\quad \ref{an:l0}\,$D_0$\qquad \ref{an:ls}\,\method\qquad \ref{an:le}\,GSA on $E(x)$};
\end{tikzpicture}
\vspace{-6pt}
\caption{\textbf{Effect of where GSA is applied.} DiT-XL/2 in the ablation setting. (a) Generation (FID on 10k samples) and reconstruction quality of all trained decoders. (b) Radially averaged power spectrum of decoded samples relative to real images.}
\label{fig:analysis}
\end{figure}
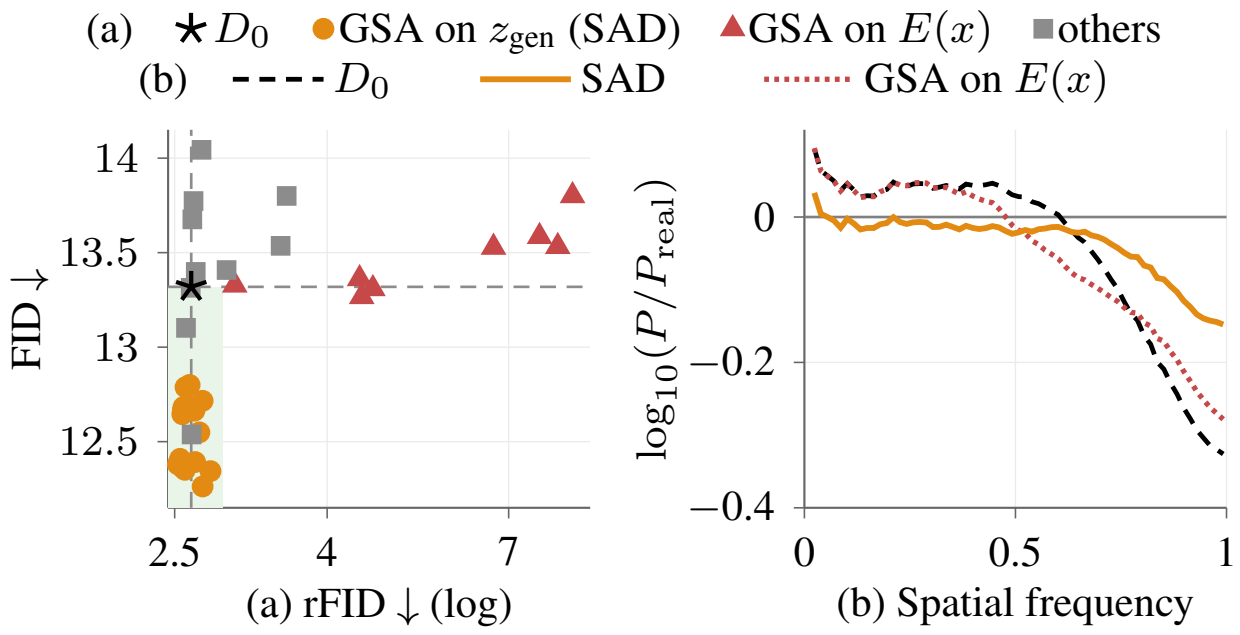

\begin{table}[t]
\centering
\footnotesize
\setlength{\tabcolsep}{5pt}
\caption{\textbf{Distribution test of generated latents.} DiT-XL/2, 10k latents each; $P(z)=E(D_0(z))$. AUC: encoder \vs generated.}
\label{tab:shift}
\vspace{2pt}
\begin{tabular}{lcc}
\toprule
\textbf{Statistic} & Encoder & Generated \\
\midrule
Relative change $\|P(z)-z\|^2/\|z\|^2$ & 0.017 & 0.016 \\
AUC of this change & -- & 0.49 \\
AUC of a small CNN on $z$ & -- & 0.96 \\
AUC of a small CNN on $P(z)$ & -- & 0.95 \\
\bottomrule
\end{tabular}
\end{table}

\noindent\textbf{Why generated latents.} \Cref{fig:analysis}a tests \cref{prop:enc,prop:opt} on every decoder we trained. All \method decoders fall in the lower-left region, better than $D_0$ in FID at unchanged rFID, whereas every decoder trained with GSA on encoder latents moves right without moving down. \Cref{tab:shift} explains why. Re-encoding changes generated latents as little as encoder latents, and a detector based on this change is at chance, whereas a small CNN separates the two distributions, equally well after re-encoding. Generated latents thus differ from encoder latents in distribution, not in their distance to the manifold. This is why robustness to perturbed latents does not help (\cref{tab:adapt}), while an alignment on the distribution of generated latents does.

\begin{figure}[t]
\centering
\begin{tikzpicture}
\begin{axis}[sadplot, scale only axis, width=0.78\linewidth, height=0.28\linewidth, grid=none, ymajorgrids,
  ymode=log, xmin=-14, xmax=29, ymin=0.25, ymax=7, ytick={0.3,1,3}, yticklabels={0.3,1,3},
  log ticks with fixed point, xtick={-10,0,10,20},
  tick label style={font=\footnotesize}, label style={font=\footnotesize},
  xlabel={Classifier logit (rank estimate of $\log r(z)$)}, xlabel style={yshift=3pt},
  ylabel={$|D_\phi(z)-D_0(z)|$ ($\times10^{-2}$)}, ylabel style={yshift=-4pt},
  legend style={font=\footnotesize, at={(0.98,0.04)}, anchor=south east, draw=none, fill=white, fill opacity=0.85, text opacity=1}]
  \addplot[cDz, densely dashed, mark=square*, mark size=1.6pt] coordinates
    {(-11.09,0.306) (-7.38,0.360) (-5.18,0.398) (-2.98,0.439) (1.04,0.498)};
  \addlegendentry{Encoder latents $E(x)$}
  \addplot[cSAD, mark=*, mark size=1.8pt] coordinates
    {(-1.14,2.14) (3.35,3.00) (6.79,3.66) (11.26,4.44) (26.37,5.23)};
  \addlegendentry{Generated latents $\zgen$}
\end{axis}
\end{tikzpicture}
\vspace{-6pt}
\caption{\textbf{Per-sample correction against the density ratio.} DiT-XL/2, 10k encoder and 10k generated latents. A classifier trained to separate generated from encoder latents scores every latent, and its logit ranks latents by $r(z)$. Each point is one quintile of a set by this score; $y$: mean pixel change $|D_\phi(z)-D_0(z)|$ made by \method.}
\label{fig:ratio}
\end{figure}
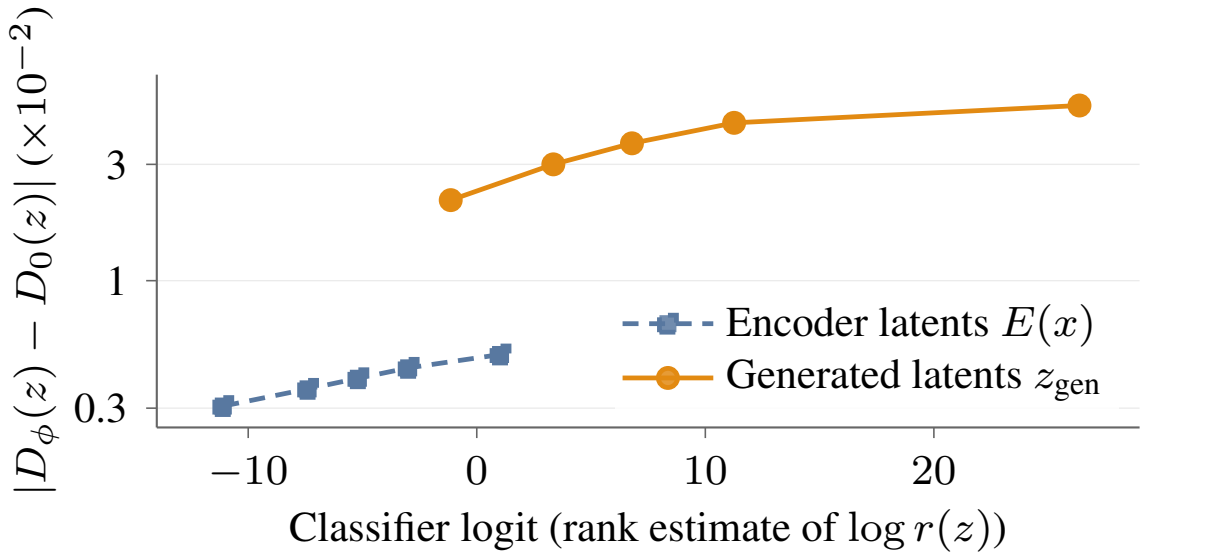

\phead{The correction follows the density ratio.} We test \cref{prop:opt} per sample by scoring every latent with a small CNN cross-fitted on 10k encoder and 10k generated latents (held-out AUC $0.95$), whose logit serves as a rank estimate of $\log r(z)$. As shown in \cref{fig:ratio}, \method changes generated latents $9{\times}$ more than encoder latents ($3.7$ \vs $0.40$ in units of $10^{-2}$), and within each set the change grows monotonically with the score (Spearman $\rho=0.57$ on generated and $0.31$ on encoder latents). The decoder thus concentrates its correction on the latents that the generator over-represents, as \cref{prop:opt} predicts.

\phead{What the decoder changes.} \Cref{fig:analysis}b shows which image statistics the correction changes. Decoded DiT samples lack high-frequency energy compared with real images; \method recovers about half of this gap, while GSA on $E(x)$ does not, and owing to EAP this costs no reconstruction (rFID $2.63\!\to\!2.56$). This matches the texture corrections in \cref{fig:qual}.

\begin{table}[t]
\centering
\footnotesize
\setlength{\tabcolsep}{5pt}
\caption{\textbf{Color and exposure statistics.} DiT-XL/2, 50-step DPM-Solver++, 3k samples and 3k held-out real images. Clipped: pixels with a channel ${\geq}250$ or all channels ${\leq}5$.}
\label{tab:color}
\vspace{2pt}
\begin{tabular}{lccc}
\toprule
\textbf{Statistic} & Real & $D_0$ & \method \\
\midrule
Saturation (HSV) & 0.323 & 0.375 & \textbf{0.324} \\
Brightness (HSV) & 0.526 & 0.539 & \textbf{0.531} \\
Colorfulness~\cite{hasler2003colorfulness} & 42.0 & 47.8 & \textbf{41.3} \\
Clipped highlights (\%) & 4.8 & 9.4 & \textbf{5.9} \\
Clipped shadows (\%) & 1.4 & 4.1 & \textbf{1.9} \\
\bottomrule
\end{tabular}
\end{table}

\phead{Color and exposure.} Decoded DiT samples are more saturated and slightly brighter than real images and clip about twice as many pixels to white or black. \method brings all these statistics close to those of real images (\cref{fig:qual,tab:color}), although no term in its objective targets color.

\phead{Transfer to unseen sampler settings.} A decoder trained with one sampler setting also improves unseen settings of the same solver. With 20 instead of 50 steps, FID drops from $13.82$ to $12.68$, and at the best guidance scale of $D_0$, FID-50K drops from $3.04$ to $2.13$ (\cref{tab:transfer,tab:guidance}).

\phead{Cost.} \method trains the 49.5M-parameter decoder (7\% of DiT-XL/2) in 0.9 GPU-hours on one NVIDIA L20, and caching the latents costs one round of sampling (about 8 GPU-hours for 40k DiT-XL/2 latents), once per generator.

\phead{Limitations.} Because the anchor bounds how far each sample can move, \method corrects texture, color and local statistics but cannot add diversity or repair wrong structure. The decoder is specific to its generator and sampler family, and a DiT decoder does not transfer to SiT, although both use SD-VAE. On text-to-image generation, a large $\lambda$ blurs some faces and adds spurious texture to smooth surfaces (\cref{fig:supp_coco}).

\section{Conclusion}
\label{sec:conclusion}

We studied the decoding step of latent generative models, where a decoder trained on encoder latents renders the residual errors of the generator into the final image.
\method fine-tunes only this decoder, with GSA aligning decoded generated latents with unpaired real images and EAP keeping the decoder valid on the latent space.
Our analysis and ablations show that the alignment must act on generated latents, where it becomes a density-ratio-weighted correction.
With about one GPU-hour of training and no inference cost, \method improves four state-of-the-art ImageNet generators across three tokenizers, complements discriminator guidance and extends to text-to-image generation.
The decoder, usually frozen once the autoencoder is trained, is thus a cheap and effective target for upgrading released generators, complementary to better latent spaces and stronger generators.
Future work will combine \method with generator-side methods that address diversity and extend it to video.

{
    \small
    \bibliographystyle{ieeenat_fullname}
    \bibliography{main}

\begin{thebibliography}{63}
\providecommand{\natexlab}[1]{#1}
\providecommand{\url}[1]{\texttt{#1}}
\expandafter\ifx\csname urlstyle\endcsname\relax
  \providecommand{\doi}[1]{doi: #1}\else
  \providecommand{\doi}{doi: \begingroup \urlstyle{rm}\Url}\fi

\bibitem[Ansari et~al.(2021)Ansari, Ang, and Soh]{ansari2021dgflow}
Abdul~Fatir Ansari, Ming~Liang Ang, and Harold Soh.
\newblock Refining deep generative models via discriminator gradient flow.
\newblock In \emph{ICLR}, 2021.

\bibitem[Arbel et~al.(2019)Arbel, Korba, Salim, and Gretton]{arbel2019mmdflow}
Michael Arbel, Anna Korba, Adil Salim, and Arthur Gretton.
\newblock Maximum mean discrepancy gradient flow.
\newblock In \emph{NeurIPS}, 2019.

\bibitem[Azadi et~al.(2019)Azadi, Olsson, Darrell, Goodfellow, and
  Odena]{azadi2019drs}
Samaneh Azadi, Catherine Olsson, Trevor Darrell, Ian Goodfellow, and Augustus
  Odena.
\newblock Discriminator rejection sampling.
\newblock In \emph{ICLR}, 2019.

\bibitem[Bai et~al.(2026)Bai, Lei, Shi, et~al.]{bai2026tgo}
Jinbin Bai, Yu Lei, Qingyu Shi, et~al.
\newblock Threshold-guided optimization for visual generative models.
\newblock In \emph{ICML}, 2026.

\bibitem[Bengio et~al.(2015)Bengio, Vinyals, Jaitly, and
  Shazeer]{bengio2015scheduled}
Samy Bengio, Oriol Vinyals, Navdeep Jaitly, and Noam Shazeer.
\newblock Scheduled sampling for sequence prediction with recurrent neural
  networks.
\newblock In \emph{NeurIPS}, 2015.

\bibitem[Chen et~al.(2025{\natexlab{a}})Chen, Han, Chen,
  et~al.]{chen2025maetok}
Hao Chen, Yujin Han, Fangyi Chen, et~al.
\newblock Masked autoencoders are effective tokenizers for diffusion models.
\newblock In \emph{ICML}, 2025{\natexlab{a}}.

\bibitem[Chen et~al.(2024)Chen, Yu, Ge, Yao, Xie, Wu, Wang, Kwok, Luo, Lu, and
  Li]{chen2024pixart}
Junsong Chen, Jincheng Yu, Chongjian Ge, Lewei Yao, Enze Xie, Yue Wu, Zhongdao
  Wang, James Kwok, Ping Luo, Huchuan Lu, and Zhenguo Li.
\newblock {PixArt-$\alpha$}: Fast training of diffusion transformer for
  photorealistic text-to-image synthesis.
\newblock In \emph{ICLR}, 2024.

\bibitem[Chen et~al.(2025{\natexlab{b}})Chen, Cai, Chen, et~al.]{chen2025dcae}
Junyu Chen, Han Cai, Junsong Chen, et~al.
\newblock Deep compression autoencoder for efficient high-resolution diffusion
  models.
\newblock In \emph{ICLR}, 2025{\natexlab{b}}.

\bibitem[Chen et~al.(2025{\natexlab{c}})Chen, Girdhar, Wang,
  et~al.]{chen2025dito}
Yinbo Chen, Rohit Girdhar, Xiaolong Wang, et~al.
\newblock Diffusion autoencoders are scalable image tokenizers.
\newblock \emph{arXiv preprint arXiv:2501.18593}, 2025{\natexlab{c}}.

\bibitem[Dhariwal and Nichol(2021)]{dhariwal2021adm}
Prafulla Dhariwal and Alexander Nichol.
\newblock Diffusion models beat {GAN}s on image synthesis.
\newblock In \emph{NeurIPS}, 2021.

\bibitem[Esser et~al.(2021)Esser, Rombach, and Ommer]{esser2021taming}
Patrick Esser, Robin Rombach, and Bj{\"o}rn Ommer.
\newblock Taming transformers for high-resolution image synthesis.
\newblock In \emph{CVPR}, 2021.

\bibitem[Fang et~al.(2026{\natexlab{a}})Fang, Shu, Xu, Mou, Kong, and
  Rudner]{fang2026gar}
Xianghong Fang, Wenjie Shu, Tongda Xu, Wenlong Mou, Dehan Kong, and Tim G.~J.
  Rudner.
\newblock Bridging reconstruction and generation: A latent distribution
  perspective on evaluation and improvement.
\newblock \emph{arXiv preprint arXiv:2609.24088}, 2026{\natexlab{a}}.

\bibitem[Fang et~al.(2026{\natexlab{b}})Fang, Zhu, Aimeur, Kong, and
  Rudner]{fang2026fdda}
Xianghong Fang, Jiajun Zhu, Esma Aimeur, Dehan Kong, and Tim G.~J. Rudner.
\newblock Efficient fr{\'e}chet distributional decoder alignment for latent
  generative models.
\newblock \emph{arXiv preprint arXiv:2610.17466}, 2026{\natexlab{b}}.

\bibitem[Gao et~al.(2023)Gao, Zhou, Cheng, and Yan]{gao2023mdtv2}
Shanghua Gao, Pan Zhou, Ming-Ming Cheng, and Shuicheng Yan.
\newblock {MDTv2}: Masked diffusion transformer is a strong image synthesizer.
\newblock \emph{arXiv preprint arXiv:2303.14389}, 2023.

\bibitem[Gretton et~al.(2012)Gretton, Borgwardt, Rasch, Sch{\"o}lkopf, and
  Smola]{gretton2012kernel}
Arthur Gretton, Karsten~M. Borgwardt, Malte~J. Rasch, Bernhard Sch{\"o}lkopf,
  and Alexander Smola.
\newblock A kernel two-sample test.
\newblock \emph{JMLR}, 13:\penalty0 723--773, 2012.

\bibitem[Hasler and S{\"u}sstrunk(2003)]{hasler2003colorfulness}
David Hasler and Sabine S{\"u}sstrunk.
\newblock Measuring colorfulness in natural images.
\newblock In \emph{Human Vision and Electronic Imaging VIII}, 2003.

\bibitem[Ho and Salimans(2021)]{ho2022cfg}
Jonathan Ho and Tim Salimans.
\newblock Classifier-free diffusion guidance.
\newblock In \emph{NeurIPS Workshop on Deep Generative Models and Downstream
  Applications}, 2021.

\bibitem[Jian et~al.(2026)Jian, Li, Zhang, Wang, Jin, Li, Tan, and
  Wang]{jian2026ranke}
Siyong Jian, Siyuan Li, Luyuan Zhang, Zedong Wang, Xin Jin, Ying Li, Cheng Tan,
  and Huan Wang.
\newblock {RankE}: End-to-end post-training for discrete text-to-image
  generation with decoder co-evolution.
\newblock In \emph{NeurIPS}, 2026.

\bibitem[Johnson and Zhang(2018)]{johnson2018cfg}
Rie Johnson and Tong Zhang.
\newblock Composite functional gradient learning of generative adversarial
  models.
\newblock In \emph{ICML}, 2018.

\bibitem[Karras et~al.(2024)Karras, Aittala, Kynk{\"a}{\"a}nniemi, Lehtinen,
  Aila, and Laine]{karras2024autoguidance}
Tero Karras, Miika Aittala, Tuomas Kynk{\"a}{\"a}nniemi, Jaakko Lehtinen, Timo
  Aila, and Samuli Laine.
\newblock Guiding a diffusion model with a bad version of itself.
\newblock In \emph{NeurIPS}, 2024.

\bibitem[Kim et~al.(2023)Kim, Kim, Kwon, Kang, and Moon]{kim2023dg}
Dongjun Kim, Yeongmin Kim, Se~Jung Kwon, Wanmo Kang, and Il-Chul Moon.
\newblock Refining generative process with discriminator guidance in
  score-based diffusion models.
\newblock In \emph{ICML}, 2023.

\bibitem[Kingma and Welling(2014)]{kingma2014vae}
Diederik~P. Kingma and Max Welling.
\newblock Auto-encoding variational {B}ayes.
\newblock In \emph{ICLR}, 2014.

\bibitem[Kynk{\"a}{\"a}nniemi et~al.(2019)Kynk{\"a}{\"a}nniemi, Karras, Laine,
  Lehtinen, and Aila]{kynkaanniemi2019pr}
Tuomas Kynk{\"a}{\"a}nniemi, Tero Karras, Samuli Laine, Jaakko Lehtinen, and
  Timo Aila.
\newblock Improved precision and recall metric for assessing generative models.
\newblock In \emph{NeurIPS}, 2019.

\bibitem[Kynk{\"a}{\"a}nniemi et~al.(2023)Kynk{\"a}{\"a}nniemi, Karras,
  Aittala, Aila, and Lehtinen]{kynkaanniemi2023role}
Tuomas Kynk{\"a}{\"a}nniemi, Tero Karras, Miika Aittala, Timo Aila, and Jaakko
  Lehtinen.
\newblock The role of {ImageNet} classes in {Fr\'echet} inception distance.
\newblock In \emph{ICLR}, 2023.

\bibitem[Kynk{\"a}{\"a}nniemi et~al.(2024)Kynk{\"a}{\"a}nniemi, Aittala,
  Karras, Laine, Aila, and Lehtinen]{kynkaanniemi2024interval}
Tuomas Kynk{\"a}{\"a}nniemi, Miika Aittala, Tero Karras, Samuli Laine, Timo
  Aila, and Jaakko Lehtinen.
\newblock Applying guidance in a limited interval improves sample and
  distribution quality in diffusion models.
\newblock In \emph{NeurIPS}, 2024.

\bibitem[Lei et~al.(2025)Lei, Bai, Shi, et~al.]{lei2025psalign}
Yu Lei, Jinbin Bai, Qingyu Shi, et~al.
\newblock Personalized safety alignment for text-to-image diffusion models.
\newblock \emph{arXiv preprint arXiv:2508.01151}, 2025.

\bibitem[Leng et~al.(2025)Leng, Singh, Hou, Xing, Xie, and
  Zheng]{leng2025repae}
Xingjian Leng, Jaskirat Singh, Yunzhong Hou, Zhenchang Xing, Saining Xie, and
  Liang Zheng.
\newblock {REPA-E}: Unlocking {VAE} for end-to-end tuning of latent diffusion
  transformers.
\newblock In \emph{ICCV}, 2025.

\bibitem[Li et~al.(2017)Li, Chang, Cheng, Yang, and P{\'o}czos]{li2017mmdgan}
Chun-Liang Li, Wei-Cheng Chang, Yu Cheng, Yiming Yang, and Barnab{\'a}s
  P{\'o}czos.
\newblock {MMD GAN}: Towards deeper understanding of moment matching network.
\newblock In \emph{NeurIPS}, 2017.

\bibitem[Li et~al.(2024{\natexlab{a}})Li, Qu, Yao, Sun, and
  Moens]{li2024alleviating}
Mingxiao Li, Tingyu Qu, Ruicong Yao, Wei Sun, and Marie-Francine Moens.
\newblock Alleviating exposure bias in diffusion models through sampling with
  shifted time steps.
\newblock In \emph{ICLR}, 2024{\natexlab{a}}.

\bibitem[Li et~al.(2024{\natexlab{b}})Li, Tian, Li, Deng, and He]{li2024mar}
Tianhong Li, Yonglong Tian, He Li, Mingyang Deng, and Kaiming He.
\newblock Autoregressive image generation without vector quantization.
\newblock In \emph{NeurIPS}, 2024{\natexlab{b}}.

\bibitem[Lu et~al.(2025)Lu, Zhou, Bao, Chen, Li, and Zhu]{lu2022dpmsolverpp}
Cheng Lu, Yuhao Zhou, Fan Bao, Jianfei Chen, Chongxuan Li, and Jun Zhu.
\newblock {DPM-Solver++}: Fast solver for guided sampling of diffusion
  probabilistic models.
\newblock \emph{Machine Intelligence Research}, 22:\penalty0 730--751, 2025.

\bibitem[Ma et~al.(2024)Ma, Goldstein, Albergo, Boffi, Vanden-Eijnden, and
  Xie]{ma2024sit}
Nanye Ma, Mark Goldstein, Michael~S. Albergo, Nicholas~M. Boffi, Eric
  Vanden-Eijnden, and Saining Xie.
\newblock {SiT}: Exploring flow and diffusion-based generative models with
  scalable interpolant transformers.
\newblock In \emph{ECCV}, 2024.

\bibitem[Mathiasen and Hvilsh{\o}j(2020)]{mathiasen2020fidbackprop}
Alexander Mathiasen and Frederik Hvilsh{\o}j.
\newblock Backpropagating through {F}r{\'e}chet inception distance.
\newblock \emph{arXiv preprint arXiv:2009.14075}, 2020.

\bibitem[Meng et~al.(2022)Meng, He, Song, Song, Wu, Zhu, and
  Ermon]{meng2022sdedit}
Chenlin Meng, Yutong He, Yang Song, Jiaming Song, Jiajun Wu, Jun-Yan Zhu, and
  Stefano Ermon.
\newblock {SDEdit}: Guided image synthesis and editing with stochastic
  differential equations.
\newblock In \emph{ICLR}, 2022.

\bibitem[M{\"u}ller(1997)]{muller1997ipm}
Alfred M{\"u}ller.
\newblock Integral probability metrics and their generating classes of
  functions.
\newblock \emph{Advances in Applied Probability}, 29\penalty0 (2):\penalty0
  429--443, 1997.

\bibitem[Ning et~al.(2023)Ning, Sangineto, Porrello, Calderara, and
  Cucchiara]{ning2023input}
Mang Ning, Enver Sangineto, Angelo Porrello, Simone Calderara, and Rita
  Cucchiara.
\newblock Input perturbation reduces exposure bias in diffusion models.
\newblock In \emph{ICML}, 2023.

\bibitem[Ning et~al.(2024)Ning, Li, Su, Salah, and
  Ertugrul]{ning2024elucidating}
Mang Ning, Mingxiao Li, Jianlin Su, Albert~Ali Salah, and Itir~Onal Ertugrul.
\newblock Elucidating the exposure bias in diffusion models.
\newblock In \emph{ICLR}, 2024.

\bibitem[{OpenAI}(2023)]{openai2023consistencydecoder}
{OpenAI}.
\newblock Consistency decoder.
\newblock \url{https://github.com/openai/consistencydecoder}, 2023.

\bibitem[Oquab et~al.(2024)Oquab, Darcet, Moutakanni, et~al.]{oquab2023dinov2}
Maxime Oquab, Timoth{\'e}e Darcet, Th{\'e}o Moutakanni, et~al.
\newblock {DINOv2}: Learning robust visual features without supervision.
\newblock \emph{TMLR}, 2024.

\bibitem[Peebles and Xie(2023)]{peebles2023dit}
William Peebles and Saining Xie.
\newblock Scalable diffusion models with transformers.
\newblock In \emph{ICCV}, 2023.

\bibitem[Podell et~al.(2024)Podell, English, Lacey, et~al.]{podell2024sdxl}
Dustin Podell, Zion English, Kyle Lacey, et~al.
\newblock {SDXL}: Improving latent diffusion models for high-resolution image
  synthesis.
\newblock In \emph{ICLR}, 2024.

\bibitem[Qiu et~al.(2025)Qiu, Li, Chen, Kuen, Xu, Gu, Luo, Raj, Lin, and
  Savvides]{qiu2025robustok}
Kai Qiu, Xiang Li, Hao Chen, Jason Kuen, Xiaohao Xu, Jiuxiang Gu, Yinyi Luo,
  Bhiksha Raj, Zhe Lin, and Marios Savvides.
\newblock Image tokenizer needs post-training.
\newblock \emph{arXiv preprint arXiv:2509.12474}, 2025.

\bibitem[Rombach et~al.(2022)Rombach, Blattmann, Lorenz, Esser, and
  Ommer]{rombach2022ldm}
Robin Rombach, Andreas Blattmann, Dominik Lorenz, Patrick Esser, and Bj{\"o}rn
  Ommer.
\newblock High-resolution image synthesis with latent diffusion models.
\newblock In \emph{CVPR}, 2022.

\bibitem[Sauer et~al.(2021)Sauer, Chitta, M{\"u}ller, and
  Geiger]{sauer2021projected}
Axel Sauer, Kashyap Chitta, Jens M{\"u}ller, and Andreas Geiger.
\newblock Projected {GAN}s converge faster.
\newblock In \emph{NeurIPS}, 2021.

\bibitem[Sauer et~al.(2022)Sauer, Schwarz, and Geiger]{sauer2022styleganxl}
Axel Sauer, Katja Schwarz, and Andreas Geiger.
\newblock {StyleGAN-XL}: Scaling {StyleGAN} to large diverse datasets.
\newblock In \emph{SIGGRAPH}, 2022.

\bibitem[Sauer et~al.(2024)Sauer, Lorenz, Blattmann, and Rombach]{sauer2023add}
Axel Sauer, Dominik Lorenz, Andreas Blattmann, and Robin Rombach.
\newblock Adversarial diffusion distillation.
\newblock In \emph{ECCV}, 2024.

\bibitem[Shi et~al.(2025)Shi, Wu, Bai, et~al.]{shi2025rectok}
Qingyu Shi, Size Wu, Jinbin Bai, et~al.
\newblock {RecTok}: Reconstruction distillation along rectified flow.
\newblock \emph{arXiv preprint arXiv:2512.13421}, 2025.

\bibitem[Skorokhodov et~al.(2025)Skorokhodov, Girish, Hu,
  et~al.]{skorokhodov2025diffusability}
Ivan Skorokhodov, Sharath Girish, Benran Hu, et~al.
\newblock Improving the diffusability of autoencoders.
\newblock In \emph{ICML}, 2025.

\bibitem[Song et~al.(2023)Song, Dhariwal, Chen, and
  Sutskever]{song2023consistency}
Yang Song, Prafulla Dhariwal, Mark Chen, and Ilya Sutskever.
\newblock Consistency models.
\newblock In \emph{ICML}, 2023.

\bibitem[Sriperumbudur et~al.(2012)Sriperumbudur, Fukumizu, Gretton,
  Sch{\"o}lkopf, and Lanckriet]{sriperumbudur2012ipm}
Bharath~K. Sriperumbudur, Kenji Fukumizu, Arthur Gretton, Bernhard
  Sch{\"o}lkopf, and Gert R.~G. Lanckriet.
\newblock On the empirical estimation of integral probability metrics.
\newblock \emph{Electronic Journal of Statistics}, 6:\penalty0 1550--1599,
  2012.

\bibitem[Stein et~al.(2023)Stein, Cresswell, Hosseinzadeh,
  et~al.]{stein2023exposing}
George Stein, Jesse~C. Cresswell, Rasa Hosseinzadeh, et~al.
\newblock Exposing flaws of generative model evaluation metrics and their
  unfair treatment of diffusion models.
\newblock In \emph{NeurIPS}, 2023.

\bibitem[Sun et~al.(2024)Sun, Jiang, Chen, et~al.]{sun2024llamagen}
Peize Sun, Yi Jiang, Shoufa Chen, et~al.
\newblock Autoregressive model beats diffusion: {L}lama for scalable image
  generation.
\newblock \emph{arXiv preprint arXiv:2406.06525}, 2024.

\bibitem[Tanaka(2019)]{tanaka2019dot}
Akinori Tanaka.
\newblock Discriminator optimal transport.
\newblock In \emph{NeurIPS}, 2019.

\bibitem[Tian et~al.(2024)Tian, Jiang, Yuan, Peng, and Wang]{tian2024var}
Keyu Tian, Yi Jiang, Zehuan Yuan, Bingyue Peng, and Liwei Wang.
\newblock Visual autoregressive modeling: Scalable image generation via
  next-scale prediction.
\newblock In \emph{NeurIPS}, 2024.

\bibitem[Wallace et~al.(2024)Wallace, Dang, Rafailov, et~al.]{wallace2024dpo}
Bram Wallace, Meihua Dang, Rafael Rafailov, et~al.
\newblock Diffusion model alignment using direct preference optimization.
\newblock In \emph{CVPR}, 2024.

\bibitem[Yao et~al.(2025)Yao, Yang, and Wang]{yao2025vavae}
Jingfeng Yao, Bin Yang, and Xinggang Wang.
\newblock Reconstruction vs. generation: Taming optimization dilemma in latent
  diffusion models.
\newblock In \emph{CVPR}, 2025.

\bibitem[Yi et~al.(2023)Yi, Zhu, and Liu]{yi2023monoflow}
Mingxuan Yi, Zhanxing Zhu, and Song Liu.
\newblock {MonoFlow}: Rethinking divergence {GANs} via the perspective of
  {Wasserstein} gradient flows.
\newblock In \emph{ICML}, 2023.

\bibitem[Yin et~al.(2024)Yin, Gharbi, Zhang, Shechtman, Durand, Freeman, and
  Park]{yin2024dmd}
Tianwei Yin, Micha{\"e}l Gharbi, Richard Zhang, Eli Shechtman, Fr{\'e}do
  Durand, William~T. Freeman, and Taesung Park.
\newblock One-step diffusion with distribution matching distillation.
\newblock In \emph{CVPR}, 2024.

\bibitem[Yu et~al.(2024)Yu, Weber, Deng, Shen, Cremers, and Chen]{yu2024titok}
Qihang Yu, Mark Weber, Xueqing Deng, Xiaohui Shen, Daniel Cremers, and
  Liang-Chieh Chen.
\newblock An image is worth 32 tokens for reconstruction and generation.
\newblock In \emph{NeurIPS}, 2024.

\bibitem[Yu et~al.(2025)Yu, Kwak, Jang, Jeong, Huang, Shin, and
  Xie]{yu2025repa}
Sihyun Yu, Sangkyung Kwak, Huiwon Jang, Jongheon Jeong, Jonathan Huang, Jinwoo
  Shin, and Saining Xie.
\newblock Representation alignment for generation: Training diffusion
  transformers is easier than you think.
\newblock In \emph{ICLR}, 2025.

\bibitem[Zhao et~al.(2025)Zhao, Woo, Wan, et~al.]{zhao2024epsvae}
Long Zhao, Sanghyun Woo, Ziyu Wan, et~al.
\newblock Epsilon-{VAE}: Denoising as visual decoding.
\newblock In \emph{ICML}, 2025.

\bibitem[Zhao et~al.(2020)Zhao, Liu, Lin, Zhu, and Han]{zhao2020diffaugment}
Shengyu Zhao, Zhijian Liu, Ji Lin, Jun-Yan Zhu, and Song Han.
\newblock Differentiable augmentation for data-efficient {GAN} training.
\newblock In \emph{NeurIPS}, 2020.

\bibitem[Zheng et~al.(2026)Zheng, Ma, Tong, and Xie]{zheng2025rae}
Boyang Zheng, Nanye Ma, Shengbang Tong, and Saining Xie.
\newblock Diffusion transformers with representation autoencoders.
\newblock In \emph{ICLR}, 2026.

\end{thebibliography}
}

\maketitlesupplementary
\appendix
\crefalias{section}{appendix}\crefname{appendix}{Appendix}{Appendices}\Crefname{appendix}{Appendix}{Appendices}
\renewcommand{\topfraction}{0.95}\renewcommand{\dbltopfraction}{0.95}\renewcommand{\textfraction}{0.05}\renewcommand{\floatpagefraction}{0.85}\renewcommand{\dblfloatpagefraction}{0.85}\setcounter{topnumber}{4}\setcounter{dbltopnumber}{3}\setcounter{totalnumber}{6}
\noindent This supplementary material is organized as follows:
\begin{itemize}
\item \cref{sec:proofs}: Proofs of the propositions in \cref{sec:method:analysis}.
\item \cref{sec:supp:impl}: Implementation details: training, generators and samplers, evaluation protocol and baselines.
\item \cref{sec:supp:results}: Additional results: decoder adaptation in the official setting, results across generators, unseen sampler settings, training dynamics, metrics unused in training, the distribution test and the per-sample density-ratio test.
\item \cref{sec:supp:t2i}: Text-to-image generation with PixArt-$\alpha$.
\item \cref{sec:supp:qual}: Additional qualitative results.
\end{itemize}

\section{Proofs}
\label{sec:proofs}
\noindent\textbf{Assumptions and notation.}
Throughout, $D_\phi = D_0 + \delta$, all latent distributions have densities, and $q$ is absolutely continuous with respect to $p_E$, so that the density ratio $r = q/p_E$ is well defined.
The distribution test (\cref{tab:shift,tab:g0}) supports this assumption: generated latents lie on the encoder manifold.
Recall that $p_D$ denotes the distribution of $D(z)$ for $z\sim q$, and $\tilde p_D$ that of $D(z)$ for $z\sim p_E$.

\phead{Proof of \cref{prop:enc}.}
If $\Lanc(\phi) = 0$, then $D_\phi(z) = D_0(z)$ for $p_E$-almost every $z$.
Two maps that agree almost everywhere under $p_E$ produce the same output distribution, so $\tilde p_{D_\phi} = \tilde p_{D_0}$.
When the distribution loss is applied to encoder latents, it is a function of $\tilde p_{D_\phi}$ only, and is therefore unchanged.
Hence $\Ldist(\phi) < \Ldist(\phi_0)$ implies $\Lanc(\phi) > 0$. \hfill$\square$

\phead{Proof of \cref{prop:opt}.}
Let $\mu_\delta$ be the distribution of the features $\psi\big((D_0+\delta)(z)\big)$, $z\sim q$, and $\nu$ that of $\psi(x)$, $x\sim p_{\mathrm{data}}$.
The squared MMD is $\mathrm{MMD}_\psi^2 = \|m_{\mu_\delta} - m_\nu\|^2_{\mathcal H}$, where $m$ denotes the kernel mean embedding.
Moving every feature $y=\psi(D_0(z))$ by a small displacement $\eta(z)$ changes it to first order by~\cite{arbel2019mmdflow}
\begin{equation*}
\mathrm{MMD}_\psi^2(\mu_\delta,\nu) - \mathrm{MMD}_\psi^2(\mu_0,\nu) \approx 2\,\mathbb E_{z\sim q}\langle \nabla w(y),\, \eta(z)\rangle,
\end{equation*}
where $w = m_{\mu_0} - m_\nu$ is the witness function.
By the chain rule, $\eta(z) \approx J_\psi\,\delta(z)$.
Rewriting the expectation over $q$ as an expectation over $p_E$ with weight $r$, the linearized objective becomes
\begin{equation*}
\mathbb E_{z\sim p_E}\big[\|\delta(z)\|^2 + 2\lambda\, r(z)\,\langle J_\psi^\top \nabla w,\, \delta(z)\rangle\big] + \text{const}.
\end{equation*}
This objective decouples over $z$.
Each term is a convex quadratic in $\delta(z)$, minimized at $\delta^\star(z) = -\lambda\, r(z)\, J_\psi^\top\nabla w$.\unskip\nobreak\hfill$\square$

For the hinge GAN loss, $\Ldist = -\mathbb E_{z\sim q}\, h\big(\psi(D_\phi(z))\big)$, and the same computation gives $\delta^\star = \tfrac{\lambda}{2}\, r\, J_\psi^\top \nabla h$: samples move up the discriminator, which plays the role of $-w$.

\phead{What a decoder cannot fix.}
The anchor also bounds how far a decoder can move the decoded distribution.
\needspace{6\baselineskip}
\begin{proposition}[What a decoder cannot fix]
\label{prop:limit}
Let $\psi$ be $L_\psi$-Lipschitz, let $\|r\|_\infty<\infty$, and let $W_2^\psi$ denote the Wasserstein-2 distance between $\psi$-features.
For every decoder in the $\ell_2$ trust region $\mathbb E_{p_E}\|D_\phi - D_0\|^2 \le \epsilon^2$,
\begin{equation}
W_2^\psi(p_{D_\phi},\, p_{\mathrm{data}}) \;\ge\; W_2^\psi(p_{D_0},\, p_{\mathrm{data}}) - L_\psi \sqrt{\|r\|_\infty}\,\epsilon .
\label{eq:limit}
\end{equation}
\end{proposition}

\phead{Proof of \cref{prop:limit}.}
Coupling $p_{D_\phi}$ and $p_{D_0}$ through the same latent $z$ gives
\begin{align*}
W_2^2(p_{D_\phi}, p_{D_0}) &\le \mathbb E_{z\sim q}\|\delta(z)\|^2 \\
&= \mathbb E_{z\sim p_E}\big[r(z)\|\delta(z)\|^2\big] \le \|r\|_\infty\,\epsilon^2 .
\end{align*}
Mapping both distributions through the $L_\psi$-Lipschitz network $\psi$ scales this distance by at most $L_\psi$, so $W_2^\psi(p_{D_\phi}, p_{D_0}) \le L_\psi\sqrt{\|r\|_\infty}\,\epsilon$.
The triangle inequality for $W_2^\psi$ then gives
$W_2^\psi(p_{D_\phi}, p_{\mathrm{data}}) \ge W_2^\psi(p_{D_0}, p_{\mathrm{data}}) - W_2^\psi(p_{D_\phi}, p_{D_0})$, which proves the claim. \hfill$\square$

\noindent The bound is loose in high dimensions, where $\|r\|_\infty$ can be large, and we use it only qualitatively: a decoder that respects the anchor can only move samples a short distance.
It can repair local errors such as texture, color statistics and fine structure, but not errors that require moving samples far, such as missing modes or the loss of diversity caused by strong guidance.
This matches our results at a guidance scale of 4.0 (\cref{tab:guidance}): the held-out \fdclip{} improves substantially, while FID, which is dominated by diversity at high guidance, barely changes.

\phead{Scope of the analysis.}
The propositions treat the decoder as a free function and use an $\ell_2$ anchor and the MMD.
In practice, the decoder is a fixed network, the anchor is $\ell_1$, and the GAN discriminator is learned.
The analysis is meant to explain the qualitative behavior of \method, namely which inputs the loss should see, where the decoder changes, and what it cannot fix, rather than to predict exact numbers.

\section{Implementation Details}
\label{sec:supp:impl}

\begin{algorithm}[t]
\caption{\textbf{Training of \method.} $\mathrm{sg}$: stop-gradient.}
\label{alg:sad}
\small
\begin{algorithmic}[1]
\Require frozen generator $G$ and sampler; frozen encoder $E$ and decoder $D_0$; frozen features $\psi$; real images $\mathcal X_r$ (unpaired); weight $\lambda$
\State $\mathcal Z \gets \{z_i\}_{i=1}^{N}$, \ $z_i \sim q$ \Comment{sample once, cache}
\State $\phi \gets \phi_0$ \Comment{initialize $D_\phi$ at $D_0$}
\For{$k = 1,\dots,K$}
  \State $x \sim p_{\mathrm{data}}$; \ $z_e \gets E(x)$
  \State $\Lanc \gets \| D_\phi(z_e) - D_0(z_e) \|_1$ \Comment{EAP}
  \State $z \sim \mathcal Z$; \ $\hat y \gets \psi(D_\phi(z))$; \ $y_r \gets \psi(x_r)$, $x_r \sim \mathcal X_r$
  \If{GAN}
    \State update heads $h$ with $\big[1-h(y_r)\big]_+ + \big[1+h(\mathrm{sg}(\hat y))\big]_+$
    \State $\Ldist \gets -h(\hat y)$
  \Else
    \State $\Ldist \gets \widehat{\mathrm{MMD}}{}^2(\hat y, y_r)$ \ or \ $\mathrm{FD}(\hat y, y_r)$
  \EndIf
  \State $\phi \gets \textsc{AdamW}\big(\phi,\ \nabla_\phi [\Lanc + \lambda \Ldist]\big)$
\EndFor
\State \Return $D_\phi$ \Comment{use $D_\phi$ in place of $D_0$}
\end{algorithmic}
\end{algorithm}

\Cref{alg:sad} summarizes the training procedure of \method.

\phead{Feature network and losses.}
All distribution losses use a frozen DINOv2 ViT-S/14~\cite{oquab2023dinov2}.
Decoded images in $[-1,1]$ are resized to $224{\times}224$ with bicubic interpolation and normalized with the ImageNet statistics.
\emph{GAN:} following projected and adversarial-distillation discriminators~\cite{sauer2021projected,sauer2023add}, we attach a light head to the patch tokens of blocks 2, 5, 8 and 11 (two 1D convolutions with 256 channels and LeakyReLU, then a linear output per token).
We use the hinge loss, and the same differentiable augmentation~\cite{zhao2020diffaugment} (random horizontal flip and translation by up to $1/8$ of the image size) is applied to real and decoded images.
The heads are trained with Adam (learning rate $2{\times}10^{-4}$, $\beta{=}(0,0.99)$), one discriminator step per decoder step.
\emph{MMD:} the unbiased estimate of $\mathrm{MMD}^2$ with a sum of five Gaussian kernels, with bandwidths $\{0.25,0.5,1,2,4\}$ times the median squared distance between real features.
\emph{FD:} the Fréchet distance between the real statistics, computed once on split A, and the statistics of the current batch together with a queue of the 1{,}024 most recent detached features.

\phead{Training.}
Each step uses 16 cached generated latents for $\Ldist$ and 8 encoder latents $E(x)$ of ImageNet training images for $\Lanc$.
The decoder is optimized with AdamW (learning rate $10^{-5}$, $\beta{=}(0.9,0.99)$, no weight decay) for 2{,}000 steps in bfloat16 autocast.
Training takes about one hour on a single GPU, most of it spent in the DINOv2 forward and backward passes.
The original decoder $D_0$ is kept frozen to compute anchor targets.

\phead{Generated latents.}
In the ablation setting, DiT-XL/2 latents for training are sampled with DPM-Solver++ (50 steps, guidance 1.5, uniform labels), with seeds that are disjoint from all evaluation seeds.
SiT-XL/2 latents use the Heun ODE solver with 50 steps, guidance on all four latent channels, and the same scaling factor $0.18215$.
For the official setting of DiT-XL/2, we use the released code and checkpoint: both the 40k training latents (seeds disjoint from evaluation) and the 50k evaluation latents use 250-step DDPM sampling with learned variance and guidance 1.5 on the first three latent channels, in fp32.
For SiT with REPA~\cite{yu2025repa}, the 16k training latents are sampled with the cheaper 50-step Heun ODE, while the 50k evaluation latents use the 250-step SDE sampler with guidance 1.8 applied in the interval of REPA; \method therefore also transfers from a deterministic to a stochastic sampler of the same model.
For LightningDiT-XL with VA-VAE~\cite{yao2025vavae}, we follow the configuration that reports FID 1.35: the Euler ODE with 250 steps, guidance 6.7 with the interval starting at 0.125, and a timestep shift of 0.3; latents have $32$ channels at $16{\times}16$, the training pool has 20k latents, and the anchor encodes ImageNet images with the VA-VAE encoder.
For PixArt-$\alpha$~\cite{chen2024pixart} at $256{\times}256$, latents are sampled with the default scheduler of the pipeline for 20 steps at guidance 4.5. The 40k training latents use COCO training captions, the 10k evaluation latents use two captions for each of the 5k val2017 images, and the real target and anchor images are 10k images from COCO test2017, which shares no image with the evaluation set.
All other hyperparameters of \method are unchanged across generators.

\phead{Discriminator guidance and REPA-E.}
DG~\cite{kim2023dg} uses the released discriminator (an ADM classifier on $32{\times}32{\times}4$ latents with a discriminator head) on top of the official DiT-XL/2 code. We follow the released sampling recipe: 250 DDPM steps, guidance 1.25 for $t>200$ and 3.0 below, and a discriminator-gradient weight of 1 for $t>100$; the evaluation seeds are those of our DiT-XL/2 run, so labels and initial noise are shared. \method is applied with the decoder trained on plain DiT-XL/2 latents, without retraining on DG samples.
For REPA-E~\cite{leng2025repae}, we use the released SiT-XL/1 trained with REPA on the frozen E2E-VAE, sampled with its recipe (250-step Euler--Maruyama SDE, guidance 2.4 on the interval $[0,0.73]$, 50 images per class). The 40k training latents use the same SDE sampler with random labels and seeds disjoint from evaluation; a decoder trained on 40k latents from the cheaper 50-step Heun ODE gives FID-50K $1.02$ (\cref{tab:generality}). The anchor and rFID use the same E2E-VAE encoder as the released model.

\phead{Evaluation protocol.}
The ADM ImageNet-256 reference batch (10k images) is split once at random into two halves: split A (5k) is the real target of $\Ldist$, and split B (5k) is the reference for all 10k-sample metrics, so no evaluation image is used in training.
Evaluation latents use seeds that are disjoint from the training latents; all decoders are evaluated on the \emph{same} latents, so differences between decoders are not affected by sampling noise.
FID uses the pytorch-fid Inception network; \fdclip{} and \fddino{} use the pooled features of CLIP ViT-B/16 and DINOv2 ViT-S/14 at $224{\times}224$; precision and recall use $k{=}3$ in Inception space~\cite{kynkaanniemi2019pr}.
rFID compares reconstructions of 10k ImageNet images (75 classes) with the originals.
FID-50K compares 50k samples with the ADM training-set statistics ($\mu,\Sigma$ from the reference batch).

\phead{Baselines.}
\emph{Noise-robust decoders} minimize $\|D_\phi(E(x)+\sigma\|E(x)\|_{\mathrm{rms}}\,\varepsilon)-D_0(E(x))\|_1$ with the anchor loss.
\emph{Paired post-training} noises a real latent to DDPM step $t_0$, denoises it with DiT-XL/2 (guidance 1.5), and trains $D_\phi$ to map the result to $D_0(E(x))$ with $\ell_1$, optionally plus our GAN loss.
\emph{Pixel refiners} are residual CNNs with three resolution levels (widths $w,2w,4w$; $w{=}64$ gives 6.8M parameters) applied to the output of the frozen $D_0$, trained with the same losses and data as \method.

\begin{table}[h]
\centering
\footnotesize
\setlength{\tabcolsep}{2.4pt}
\caption{\textbf{Training and inference cost of \method.}}
\label{tab:supp:cost}
\vspace{3pt}
\begin{tabular}{lcccc}
\toprule
 & \makecell{trainable\\params} & \makecell{extra params\\at inference} & \makecell{extra GFLOPs\\per image} & \makecell{training\\(GPU-h)} \\
\midrule
\rowcolor{oursbg} \method & 49.5M & 0 & 0 & 0.9 \\
Pixel refiner & 6.8M & 6.8M & 95 ($+15\%$) & 0.6 \\
\bottomrule
\end{tabular}
\end{table}

\phead{Cost.}
\Cref{tab:supp:cost} compares \method with a pixel refiner trained with the same objective: \method adds no parameters or computation at inference, whereas the refiner adds 15\% to the decoding cost.

\section{Additional Results}
\label{sec:supp:results}

\begin{table}[h]
\centering
\footnotesize
\setlength{\tabcolsep}{3.5pt}
\caption{\textbf{Comparison with decoder adaptation methods in the official setting.} DiT-XL/2 at guidance 1.5 (FID-50K, ADM evaluator; \fdclip{} with a held-out reference). All decoders are trained on the same 40k generated latents. The best results are in bold and the second best are underlined. $^\ast$Inception, the FID network, is part of its training objective.}
\label{tab:adapt_official}
\vspace{3pt}
\begin{tabular}{lcccc}
\toprule
\textbf{Method} & FID$\downarrow$ & sFID$\downarrow$ & IS$\uparrow$ & \fdclip$\downarrow$ \\
\midrule
Original decoder $D_0$ & 2.30 & 4.51 & 277.8 & 42.8 \\
Noise-robust decoder~\cite{zheng2025rae} & 2.31 & 4.64 & 277.0 & 43.1 \\
Paired post-training~\cite{qiu2025robustok,fang2026gar} & 2.69 & 7.67 & 266.7 & 54.9 \\
FDDA~\cite{fang2026fdda}$^\ast$ & \textbf{2.02} & 4.84 & 274.9 & 48.9 \\
FDDA~\cite{fang2026fdda}$^\ast$ + EAP & 2.27 & \textbf{4.14} & 279.4 & 51.2 \\
\midrule
\rowcolor{oursbg} \method-GAN (ours) & \underline{2.13} & \underline{4.25} & \underline{280.5} & \textbf{36.8} \\
\rowcolor{oursbg} \method-MMD (ours) & 2.17 & 4.36 & \textbf{280.6} & \underline{39.4} \\
 
\bottomrule
\end{tabular}
\end{table}

\noindent\textbf{Decoder adaptation in the official setting.} Does the ranking of \cref{tab:adapt} hold with FID-50K and the official DiT sampler? \Cref{tab:adapt_official} shows that it does. \method improves every metric over $D_0$ and gives by far the best held-out \fdclip{} ($42.8\!\to\!36.8$), while noise-robust and paired post-training do not improve $D_0$. FDDA, which optimizes Inception features, reaches the lowest FID but worsens sFID, IS and \fdclip{} and doubles rFID ($2.63\!\to\!5.46$); adding our EAP to it restores rFID ($2.65$) but removes most of its FID gain ($2.27$), while \fdclip{} remains worse than $D_0$. Optimizing the network of the evaluation metric thus lowers FID without improving the samples, which is why \method aligns in a separate feature space.

\begin{table}[h]
\centering
\footnotesize
\setlength{\tabcolsep}{2pt}
\renewcommand{\arraystretch}{1.06}
\caption{\textbf{Results across generators and datasets.} Each decoder is trained on latents of its own generator. ODE pool: training latents from the 50-step Heun ODE instead of the deployed SDE sampler. Pick: share of samples for which PickScore prefers \method over $D_0$ (ImageNet prompts: ``a photo of a \textit{class}'').}
\label{tab:generality}
\vspace{3pt}
\resizebox{\linewidth}{!}{%
\begin{tabular}{lcccc}
\toprule
\textbf{$D_0\!\to$ \method} & \hdr{FID$\downarrow$} & \hdr{\fdclip$\downarrow$} & \hdr{rFID$\downarrow$} & \hdr{Pick} \\
\midrule
\multicolumn{5}{l}{\textit{DiT-XL/2 on ImageNet (50k samples)}} \\
DPM-50, GAN & 3.24$\to$2.28 & 48.3$\to$39.4 & 2.63$\to$2.56 & 51\% \\
DDPM-250 (official), GAN & 2.30$\to$2.13 & 42.8$\to$36.8 & 2.63$\to$2.69 & 50\% \\
DDPM-250 (official), MMD & 2.30$\to$2.17 & 42.8$\to$39.4 & 2.63$\to$2.66 & 52\% \\
\midrule
\multicolumn{5}{l}{\textit{LightningDiT with VA-VAE on ImageNet (50k samples)}} \\
LightningDiT-XL, GAN & 1.38$\to$1.23 & 29.4$\to$28.2 & 1.18$\to$1.17 & 70\% \\
LightningDiT-XL, MMD & 1.38$\to$1.29 & 29.4$\to$27.2 & 1.18$\to$1.21 & 61\% \\
\midrule
\multicolumn{5}{l}{\textit{SiT on ImageNet}} \\
SiT-XL/2, GAN (10k) & 12.88$\to$12.80 & 47.2$\to$42.6 & 2.63$\to$2.71 & 61\% \\
SiT+REPA, GAN (50k) & 1.43$\to$1.38 & 39.9$\to$39.4 & 2.63$\to$2.59 & 58\% \\
SiT+REPA, MMD (50k) & 1.43$\to$1.36 & 39.9$\to$35.9 & 2.63$\to$2.63 & 46\% \\
\midrule
\multicolumn{5}{l}{\textit{REPA-E with E2E-VAE on ImageNet (50k samples)}} \\
SiT-XL/1, GAN & 1.15$\to$0.98 & 29.1$\to$26.4 & 2.00$\to$1.87 & 55\% \\
SiT-XL/1, GAN, ODE pool & 1.15$\to$1.02 & 29.1$\to$26.5 & 2.00$\to$1.78 & 61\% \\
SiT-XL/1, MMD, ODE pool & 1.15$\to$1.08 & 29.1$\to$26.6 & 2.00$\to$1.95 & 54\% \\
\midrule
\multicolumn{5}{l}{\textit{PixArt-$\alpha$ on COCO (10k samples)}} \\
GAN, $\lambda{=}0.01$ & 34.73$\to$23.55 & 209$\to$142 & 4.28$\to$4.29 & 48\% \\
GAN, $\lambda{=}0.05$ & 34.73$\to$20.45 & 209$\to$117 & 4.28$\to$4.91 & 24\% \\
MMD, $\lambda{=}0.1$ & 34.73$\to$28.35 & 209$\to$150 & 4.28$\to$4.26 & 67\% \\
MMD, $\lambda{=}0.3$ & 34.73$\to$26.19 & 209$\to$125 & 4.28$\to$4.31 & 50\% \\
 
\bottomrule
\end{tabular}}
\end{table}

\phead{Results across generators.} Does the gain extend beyond FID and the generators of \cref{tab:sota}? \Cref{tab:generality} adds \fdclip{} and reconstruction for every generator, the ablation setting, SiT-XL/2 without REPA and PixArt-$\alpha$. \method lowers \fdclip{} in every setting at unchanged reconstruction, so the gain is specific neither to one generator nor to the Inception space.

\begin{table}[h]
\centering
\small
\setlength{\tabcolsep}{4pt}
\renewcommand{\arraystretch}{1.1}
\caption{\textbf{Transfer of one decoder to other sampler settings.} \method values are means over three training seeds.}
\label{tab:transfer}
\vspace{3pt}
\begin{tabular}{lrrrr}
\toprule
\multirow{2}{*}{\textbf{Sampling setting}} & \multicolumn{2}{c}{\textbf{FID}$\downarrow$} & \multicolumn{2}{c}{\textbf{\fdclip}$\downarrow$} \\
\cmidrule(lr){2-3} \cmidrule(lr){4-5}
 & \hdr{$D_0$} & \hdr{\method} & \hdr{$D_0$} & \hdr{\method} \\
\midrule
cfg 1.5, 50 steps (train) & 13.32 & 12.37\,\gain{$-$7\%} & 51.4 & 44.2\,\gain{$-$14\%} \\
cfg 1.5, 20 steps & 13.82 & 12.68\,\gain{$-$8\%} & 52.8 & 42.0\,\gain{$-$21\%} \\
cfg 1.0, 50 steps & 18.68 & 18.39\,\gain{$-$2\%} & 68.8 & 68.5\,\gain{$-$1\%} \\
cfg 4.0, 50 steps & 26.50 & 25.75\,\gain{$-$3\%} & 70.2 & 49.5\,\gain{$-$29\%} \\
 
\bottomrule
\end{tabular}
\end{table}
\begin{table}[h]
\centering
\small
\setlength{\tabcolsep}{4.2pt}
\renewcommand{\arraystretch}{1.1}
\caption{\textbf{\method combined with guidance tuning.} \method values are means over three training seeds.}
\label{tab:guidance}
\vspace{3pt}
\begin{tabular}{ccrrrr}
\toprule
\multirow{2}{*}{\textbf{Scale}} & \multirow{2}{*}{\textbf{Interval}} & \multicolumn{2}{c}{\textbf{FID}$\downarrow$} & \multicolumn{2}{c}{\textbf{\fdclip}$\downarrow$} \\
\cmidrule(lr){3-4} \cmidrule(lr){5-6}
 & & \hdr{$D_0$} & \hdr{\method} & \hdr{$D_0$} & \hdr{\method} \\
\midrule
1.5 & -- & 13.32 & \textbf{12.37}\,\gain{$-$7\%} & 51.4 & 44.2\,\gain{$-$14\%} \\
1.75 & -- & 15.31 & 14.34\,\gain{$-$6\%} & 49.9 & 40.1\,\gain{$-$20\%} \\
2.0 & \checkmark & 13.53 & 12.65\,\gain{$-$6\%} & 49.6 & 40.7\,\gain{$-$18\%} \\
2.5 & \checkmark & 14.98 & 14.19\,\gain{$-$5\%} & 50.1 & 38.4\,\gain{$-$23\%} \\
3.0 & \checkmark & 16.51 & 15.82\,\gain{$-$4\%} & 51.9 & 37.9\,\gain{$-$27\%} \\
 
\bottomrule
\end{tabular}
\end{table}

\phead{Unseen sampler settings.} Must the decoder be retrained whenever the sampler setting changes? Not within a solver: a single decoder, trained at 50 steps and guidance 1.5, improves sampler settings it has never seen (\cref{tab:transfer,tab:guidance}): with 20 steps the gain is fully preserved, and across guidance scales, with or without the guidance interval of~\cite{kynkaanniemi2024interval}, \method lowers FID by 4--7\% and \fdclip{} by 14--27\%. On a finer FID-50K grid, $D_0$ is best at a scale of 1.375 ($3.04$) and \method, trained only at 1.5, reduces it to $2.13$, so the gain stacks with guidance tuning. At strong guidance the FID gain is small although \fdclip{} drops sharply, because FID is then dominated by the loss of diversity, which a decoder cannot restore (\cref{prop:limit}).
\phead{Specificity to the generator and solver.} Across solver families, however, the decoder must be retrained. With the diffusers implementation of the DiT-XL/2 checkpoint, a decoder trained on 50-step DPM-Solver++ latents does not transfer to 250-step DDPM sampling with three-channel guidance (FID-50K $2.76\!\to\!2.82$, sFID $5.32\!\to\!6.04$), whereas a decoder trained on latents of that sampler lowers FID-50K to $2.28$ (sFID $4.87$, IS $270.8\!\to\!276.0$). The correction is likewise specific to the generator: a decoder trained on DiT-XL/2 latents raises \fdclip{} on SiT-XL/2 samples from $47.2$ to $59.4$. \method thus corrects the residual errors of the deployed generator and sampler, and one decoder is trained for each.

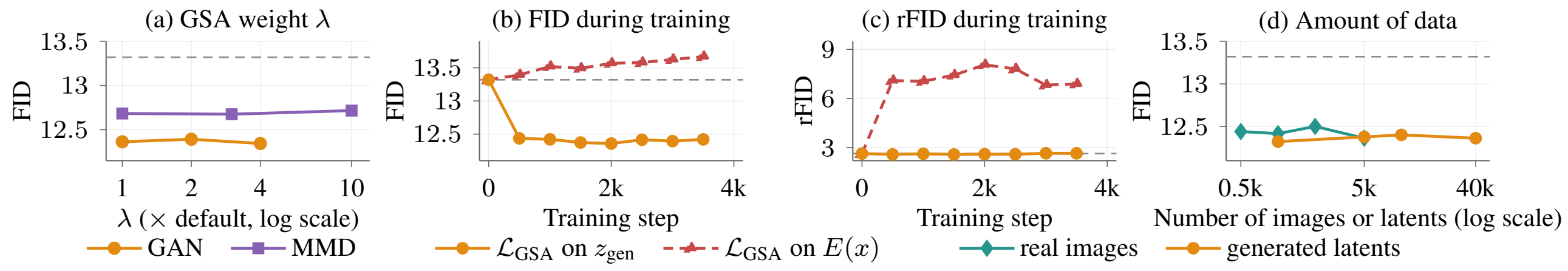
\begin{figure*}[t]
\centering
\begin{tikzpicture}
\pgfplotsset{abl/.style={sadplot, width=0.255\textwidth, height=0.165\textwidth, tick label style={font=\footnotesize},
  xlabel style={yshift=3pt, font=\footnotesize}, ylabel style={yshift=-4pt, font=\footnotesize}, title style={font=\footnotesize, yshift=-3pt}}}
\begin{axis}[abl, name=A, xmode=log, xmin=0.85, xmax=12, ymin=12.15, ymax=13.5, ytick={12.5,13,13.5},
  xtick={1,2,4,10}, xticklabels={1,2,4,10}, log ticks with fixed point, minor xtick={},
  xlabel={$\lambda$ ($\times$ default, log scale)}, ylabel={FID}, title={(a) GSA weight $\lambda$}]
  \draw[black!45, densely dashed, line width=0.5pt] (axis cs:0.85,13.319) -- (axis cs:12,13.319);
  \addplot[cSAD, mark=*, mark size=1.6pt] table[x expr=\thisrow{lam}/0.05, y=fid] {fig/data/abl_lam_gan.dat};\label{ab:lgan}
  \addplot[cRef, mark=square*, mark size=1.4pt] table[x=lam, y=fid] {fig/data/abl_lam_mmd.dat};\label{ab:lmmd}
\end{axis}
\begin{axis}[abl, name=B, at={(A.south east)}, anchor=south west, xshift=34pt, xmin=-150, xmax=4150, ymin=12.1, ymax=13.9, ytick={12.5,13,13.5},
  xtick={0,2000,4000}, xticklabels={0,2k,4k}, xlabel={Training step}, ylabel={FID}, title={(b) FID during training}]
  \draw[black!45, densely dashed, line width=0.5pt] (axis cs:-150,13.319) -- (axis cs:4150,13.319);
  \addplot[cEnc, densely dashed, mark=triangle*, mark size=1.8pt] table[x=step, y=fid] {fig/data/abl_dyn_enc.dat};\label{ab:enc}
  \addplot[cSAD, mark=*, mark size=1.5pt] table[x=step, y=fid] {fig/data/abl_dyn_gen.dat};\label{ab:gen}
\end{axis}
\begin{axis}[abl, name=C, at={(B.south east)}, anchor=south west, xshift=34pt, xmin=-150, xmax=4150, ymin=2.2, ymax=9.5, ytick={3,6,9},
  xtick={0,2000,4000}, xticklabels={0,2k,4k}, xlabel={Training step}, ylabel={rFID}, title={(c) rFID during training}]
  \draw[black!45, densely dashed, line width=0.5pt] (axis cs:-150,2.631) -- (axis cs:4150,2.631);
  \addplot[cEnc, densely dashed, mark=triangle*, mark size=1.8pt] table[x=step, y=rfid] {fig/data/abl_dyn_enc.dat};
  \addplot[cSAD, mark=*, mark size=1.5pt] table[x=step, y=rfid] {fig/data/abl_dyn_gen.dat};
\end{axis}
\begin{axis}[abl, name=Dd, at={(C.south east)}, anchor=south west, xshift=34pt, xmode=log, xmin=380, xmax=52000, ymin=12.1, ymax=13.5, ytick={12.5,13,13.5},
  xtick={500,5000,40000}, xticklabels={0.5k,5k,40k}, log ticks with fixed point, minor xtick={},
  xlabel={Number of images or latents (log scale)}, ylabel={FID}, title={(d) Amount of data}]
  \draw[black!45, densely dashed, line width=0.5pt] (axis cs:380,13.319) -- (axis cs:52000,13.319);
  \addplot[cPair, mark=diamond*, mark size=2pt] table[x=n, y=fid] {fig/data/abl_data_real.dat};\label{ab:real}
  \addplot[cSAD, mark=*, mark size=1.5pt] table[x=n, y=fid] {fig/data/abl_data_gen.dat};\label{ab:ngen}
\end{axis}
\node[font=\footnotesize, anchor=north] at ([yshift=-20pt]B.south east) {%
  \ref{ab:lgan}\,GAN\quad \ref{ab:lmmd}\,MMD\qquad\quad \ref{ab:gen}\,$\Ldist$ on $\zgen$\quad \ref{ab:enc}\,$\Ldist$ on $E(x)$\qquad\quad \ref{ab:real}\,real images\quad \ref{ab:ngen}\,generated latents};
\end{tikzpicture}
\vspace{-4pt}
\caption{\textbf{Training dynamics and data efficiency.} DiT-XL/2 in the ablation setting. (a) GSA weight $\lambda$ relative to the default. (b, c) FID and rFID during training. (d) Number of cached latents or real images.}
\label{fig:ablation}
\end{figure*}
\phead{Training dynamics and data.} How much training and data does \method need? \Cref{fig:ablation} varies $\lambda$, the training length and the amount of data. FID drops within 500 steps and stays flat, while GSA on encoder latents never improves FID and steadily raises rFID. 1k cached latents or 500 real images already give FID $12.3$--$12.5$, so \method needs neither long training nor large data.

\begin{table*}[t]
\centering
\footnotesize
\setlength{\tabcolsep}{8pt}
\caption{\textbf{Evaluation with metrics unused in training.} KID $\times10^3$, FD in SigLIP space, CMMD, HPS v2.1 and PickScore. Win: share of samples for which the preference model prefers the decoder (or the real image with the same caption) over $D_0$.}
\label{tab:extra_metrics}
\vspace{3pt}
\begin{tabular}{lrrrrrrr}
\toprule
\textbf{Decoder} & \hdr{KID$\downarrow$} & \hdr{FD$_\text{SigLIP}\downarrow$} & \hdr{CMMD$\downarrow$} & \hdr{HPS} & \hdr{PickScore} & \hdr{HPS win} & \hdr{Pick win} \\
\midrule
\multicolumn{8}{l}{\textit{DiT-XL/2, DPM-Solver++ (50 steps)}} \\
$D_0$ & 1.17 & 10.0 & 0.227 & 0.235 & 20.42 & -- & -- \\
$\Ldist$ on $E(x)$, GAN & 1.25 & 9.6 & 0.267 & 0.231 & 20.39 & 25\% & 45\% \\
\rowcolor{oursbg} \method-GAN & 0.59 & 8.0 & 0.267 & 0.231 & 20.42 & 31\% & 51\% \\
\rowcolor{oursbg} \method-MMD & 0.72 & 7.9 & 0.187 & 0.235 & 20.42 & 48\% & 50\% \\
\midrule
\multicolumn{8}{l}{\textit{DiT-XL/2, official code (DDPM-250)}} \\
$D_0$ & 0.64 & 7.4 & 0.283 & 0.232 & 20.40 & -- & -- \\
$\Ldist$ on $E(x)$, GAN & 1.08 & 8.5 & 0.373 & 0.229 & 20.34 & 32\% & 40\% \\
\rowcolor{oursbg} \method-GAN & 0.61 & 6.9 & 0.224 & 0.232 & 20.39 & 49\% & 50\% \\
\rowcolor{oursbg} \method-MMD & 0.64 & 8.9 & 0.239 & 0.233 & 20.40 & 61\% & 52\% \\
\midrule
\multicolumn{8}{l}{\textit{LightningDiT-XL with VA-VAE}} \\
$D_0$ & 0.25 & 5.9 & 0.143 & 0.230 & 20.46 & -- & -- \\
$\Ldist$ on $E(x)$, GAN & 0.93 & 7.0 & 0.217 & 0.231 & 20.46 & 54\% & 54\% \\
\rowcolor{oursbg} \method-GAN & 0.18 & 5.5 & 0.136 & 0.231 & 20.51 & 52\% & 70\% \\
\rowcolor{oursbg} \method-MMD & 0.24 & 5.7 & 0.136 & 0.232 & 20.48 & 66\% & 61\% \\
\midrule
\multicolumn{8}{l}{\textit{REPA-E with E2E-VAE}} \\
$D_0$ & 0.26 & 5.7 & 0.110 & 0.234 & 20.55 & -- & -- \\
\rowcolor{oursbg} \method-GAN & 0.12 & 5.0 & 0.124 & 0.233 & 20.57 & 35\% & 55\% \\
\rowcolor{oursbg} \method-GAN, ODE pool & 0.15 & 5.1 & 0.121 & 0.234 & 20.58 & 40\% & 61\% \\
\rowcolor{oursbg} \method-MMD, ODE pool & 0.22 & 5.6 & 0.096 & 0.233 & 20.56 & 30\% & 54\% \\
\midrule
\multicolumn{8}{l}{\textit{PixArt-$\alpha$ on COCO}} \\
$D_0$ & 11.21 & 43.8 & 0.745 & 0.285 & 22.17 & -- & -- \\
$\Ldist$ on $E(x)$, GAN & 10.64 & 42.8 & 0.733 & 0.286 & 22.19 & 58\% & 65\% \\
\rowcolor{oursbg} \method-GAN, $\lambda{=}0.01$ & 3.17 & 25.5 & 0.666 & 0.277 & 21.97 & 40\% & 48\% \\
\rowcolor{oursbg} \method-GAN, $\lambda{=}0.05$ & 1.99 & 18.5 & 0.729 & 0.264 & 21.64 & 16\% & 24\% \\
\rowcolor{oursbg} \method-MMD, $\lambda{=}0.1$ & 6.45 & 34.1 & 0.582 & 0.285 & 22.21 & 51\% & 67\% \\
\rowcolor{oursbg} \method-MMD, $\lambda{=}1$ & 3.83 & 24.1 & 0.665 & 0.271 & 21.89 & 12\% & 28\% \\
DiT decoder, transferred & 6.74 & 39.6 & 0.767 & 0.287 & 22.29 & 59\% & 74\% \\
Real images (val2017) & -- & -- & -- & 0.247 & 21.13 & 15\% & 18\% \\
 
\bottomrule
\end{tabular}
\end{table*}

\phead{Metrics unused in training.} Is the gain specific to the metrics in the main text? \Cref{tab:extra_metrics} lists all additional metrics, including those that do not favor \method. KID and FD$_\text{SigLIP}$ improve with the GAN discrepancy for every generator. CMMD improves for DiT-XL/2, LightningDiT and PixArt-$\alpha$, and with GAN it worsens on REPA-E ($0.110\!\to\!0.124$) and on DiT-XL/2 in the ablation setting, where MMD improves it. The two preference models rate real COCO images lower than $D_0$ samples (real images win only 15\% and 18\% of the pairs with the same caption), so they reward the look of generated images and do not measure realism alone.

\begin{table}[t]
\centering
\footnotesize
\setlength{\tabcolsep}{3pt}
\caption{\textbf{Full distribution test of generated latents.} DiT-XL/2 in the ablation setting, 10k latents per set, generated latents at guidance 1.0 and 1.5. $P(z)=E(D_0(z))$ with the posterior mean. AUC: encoder \vs generated latents.}
\label{tab:g0}
\vspace{3pt}
\begin{tabular}{lccc}
\toprule
\textbf{Statistic} & Encoder & \makecell{Generated\\(1.0)} & \makecell{Generated\\(1.5)} \\
\midrule
Relative change $\|P(z)-z\|^2/\|z\|^2$ & 0.0168 & 0.0160 & 0.0162 \\
AUC of this change & -- & 0.48 & 0.49 \\
FID of $D_0(z)$ (encoder: rFID) & 2.63 & 18.62 & 12.48 \\
FID of $D_0(P(z))$ (encoder: rFID) & 3.91 & 20.67 & 13.40 \\
\midrule
AUC of a small CNN on $z$ & -- & 0.94 & 0.96 \\
AUC of a small CNN on $P(z)$ & -- & 0.93 & 0.95 \\
AUC of a small CNN on $z-P(z)$ & -- & 0.84 & 0.86 \\
\bottomrule
\end{tabular}
\end{table}

\phead{Distribution test.}
\label{sec:supp:offmanifold}
Do generated latents lie off the encoder manifold, so that re-encoding them would help? \Cref{tab:g0} extends \cref{tab:shift} of the main text and answers no. The re-encoding $P$ changes generated latents as much as encoder latents ($0.016$ \vs $0.017$ relative change), and a detector based on this change is at chance (AUC $0.48$), whereas a small CNN separates the two sets easily (AUC $0.96$), equally well after re-encoding ($0.95$). Re-encoding also worsens the FID of the decoded samples.
A CNN that sees only the residual $z-P(z)$ still reaches AUC $0.84$--$0.86$, and per-channel standard deviations of generated latents exceed those of encoder latents by 5--20\%. The two distributions therefore differ, but not in their distance to the encoder manifold.

\phead{Per-sample density-ratio test.}
\Cref{fig:ratio} uses the 10k encoder latents and the 10k generated latents at guidance 1.5 of \cref{tab:g0}. The classifier is the small CNN of the distribution test (five convolutions with GELU, global pooling and a linear output), trained with two-fold cross-fitting so that every latent is scored by a model that has not seen it; its held-out AUC is $0.95$. The change of a sample is the mean absolute difference between $D_\phi(z)$ and $D_0(z)$ in $[0,1]$ pixel units, for the default GAN decoder. Over all 20k latents, the Spearman correlation between the logit and the change is $0.80$. Because the logits of a confident classifier are not calibrated, we use them only to rank latents by $r(z)$.

\section{Text-to-Image Generation}
\label{sec:supp:t2i}
\figsuppcoco
\begin{table}[t]
\centering
\footnotesize
\setlength{\tabcolsep}{3pt}
\caption{\textbf{\method on PixArt-$\alpha$ for all tested $\lambda$.} COCO, 10k samples. Pick win: share of samples preferred by PickScore over $D_0$ (same latent; for real val2017 images, same caption).}
\label{tab:coco_lam}
\vspace{3pt}
\begin{tabular}{lrrrrrr}
\toprule
\textbf{Decoder} & \hdr{FID$\downarrow$} & \hdr{\fdclip$\downarrow$} & \hdr{CLIP Sc.} & \hdr{Rec.} & \hdr{rFID$\downarrow$} & \hdr{Pick win} \\
\midrule
$D_0$ & 34.73 & 209.1 & 26.03 & 0.286 & 4.28 & -- \\
GAN, $\lambda{=}0.005$ & 24.67 & 155.9 & 27.21 & 0.400 & 4.28 & 39\% \\
GAN, $\lambda{=}0.01$ & 23.55 & 141.6 & 26.91 & 0.416 & 4.29 & 48\% \\
GAN, $\lambda{=}0.02$ & 25.51 & 135.6 & 26.40 & 0.365 & 4.40 & 48\% \\
GAN, $\lambda{=}0.05$ & 20.45 & 116.8 & 26.82 & 0.408 & 4.91 & 24\% \\
MMD, $\lambda{=}0.1$ & 28.35 & 150.4 & 26.19 & 0.343 & 4.26 & 67\% \\
MMD, $\lambda{=}0.3$ & 26.19 & 125.1 & 26.66 & 0.364 & 4.31 & 50\% \\
MMD, $\lambda{=}1$ & 24.24 & 114.8 & 26.76 & 0.393 & 4.97 & 28\% \\
 
\bottomrule
\end{tabular}
\end{table}

How strongly should the decoder be adapted for text-to-image generation? \Cref{tab:coco_lam} lists all values of $\lambda$ we tested on PixArt-$\alpha$.
Every setting lowers FID substantially, from $34.73$ to between $20.45$ and $28.35$, and raises recall from $0.286$ to between $0.343$ and $0.416$.
The weight $\lambda$ controls how far the decoder moves.
With GAN at $\lambda{=}0.01$ or MMD at $\lambda{=}0.1$--$0.3$, rFID stays within $0.03$ of $D_0$ and PickScore prefers \method in 48--67\% of the samples; with the largest $\lambda$, FID is lowest but rFID rises to about $4.9$ and preference drops to 24--28\%.
The relationship between FID and $\lambda$ is not monotonic for GAN, so we report the full curve.

\Cref{fig:supp_coco} shows samples from the same latents and prompts. Both values of $\lambda$ add texture to fur, grass and food and slightly reduce the saturated colors that are typical of PixArt-$\alpha$; the larger $\lambda$ changes the image more. The last two examples in the bottom row show the main failure modes at the larger $\lambda$: some faces become blurrier, and smooth surfaces such as screens receive spurious texture. We regard the drop in preference at large $\lambda$ as a limitation, and use a small $\lambda$ for text-to-image models.

\makeatletter\setlength{\@dblfptop}{0pt}\setlength{\@fptop}{0pt}\setlength{\@dblfpsep}{14pt}\setlength{\@dblfpbot}{0pt plus 1fil}\makeatother
\section{Qualitative Results}
\label{sec:supp:qual}
\begin{figure*}[t]
\centering
\setlength{\tabcolsep}{0.5pt}
\renewcommand{\arraystretch}{0}
\def\ow{1.84cm}
\begin{tabular}{@{}c@{\hspace{2pt}}ccc@{\hspace{4pt}}ccc@{\hspace{4pt}}ccc@{}}
 & \multicolumn{3}{c}{\scriptsize\bfseries\itshape (a) LightningDiT-XL with VA-VAE} & \multicolumn{3}{c}{\scriptsize\bfseries\itshape (b) REPA-E with E2E-VAE} & \multicolumn{3}{c}{\scriptsize\bfseries\itshape (c) PixArt-$\alpha$ text-to-image} \\[3pt]
\rotatebox{90}{\makebox[\ow][c]{\scriptsize $D_0$}} & \includegraphics[width=\ow]{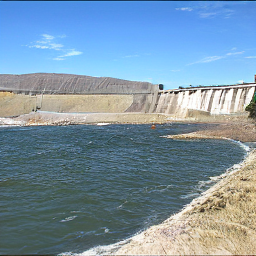} & \includegraphics[width=\ow]{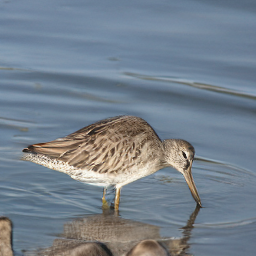} & \includegraphics[width=\ow]{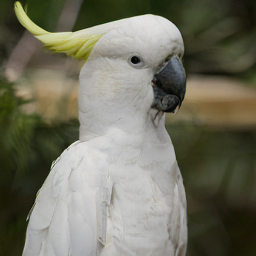} & \includegraphics[width=\ow]{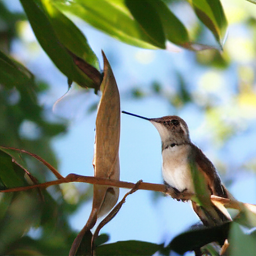} & \includegraphics[width=\ow]{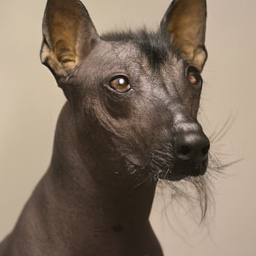} & \includegraphics[width=\ow]{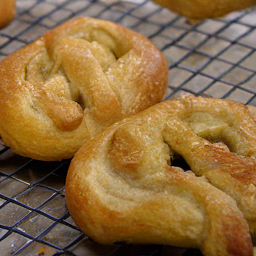} & \includegraphics[width=\ow]{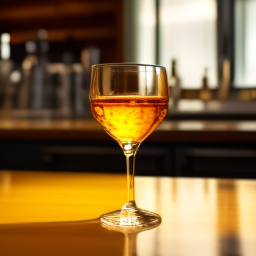} & \includegraphics[width=\ow]{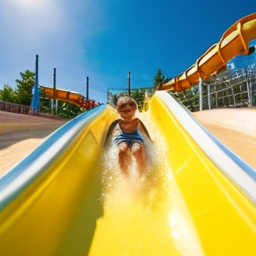} & \includegraphics[width=\ow]{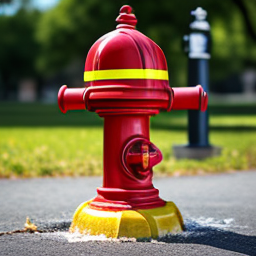} \\[1pt]
\rotatebox{90}{\makebox[\ow][c]{\scriptsize \textbf{\method}}} & \includegraphics[width=\ow]{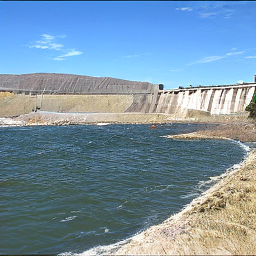} & \includegraphics[width=\ow]{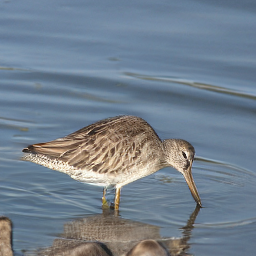} & \includegraphics[width=\ow]{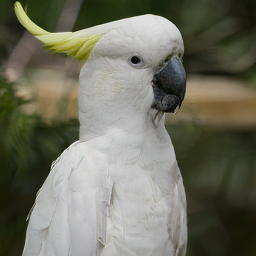} & \includegraphics[width=\ow]{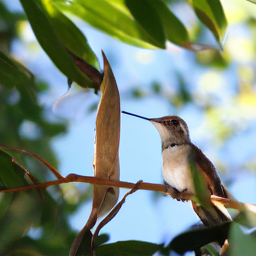} & \includegraphics[width=\ow]{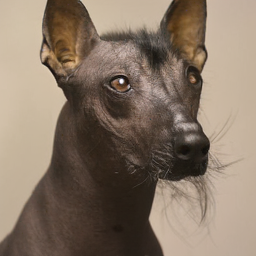} & \includegraphics[width=\ow]{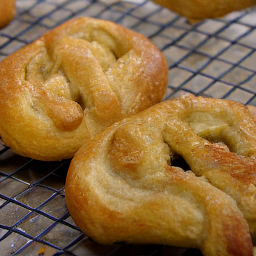} & \includegraphics[width=\ow]{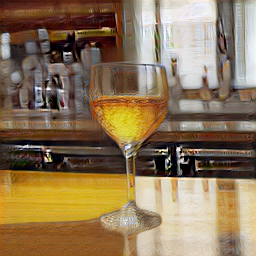} & \includegraphics[width=\ow]{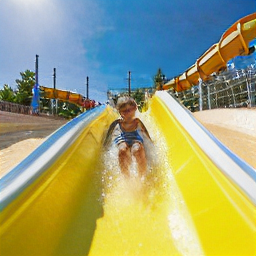} & \includegraphics[width=\ow]{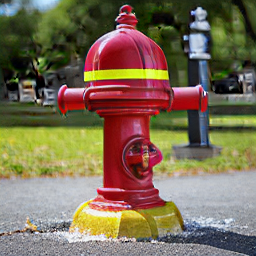} \\
\rotatebox{90}{\makebox[\ow][c]{\scriptsize $|\text{\method}-D_0|$}} & \includegraphics[width=\ow]{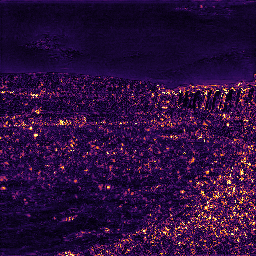} & \includegraphics[width=\ow]{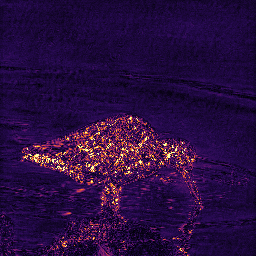} & \includegraphics[width=\ow]{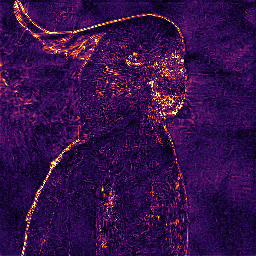} & \includegraphics[width=\ow]{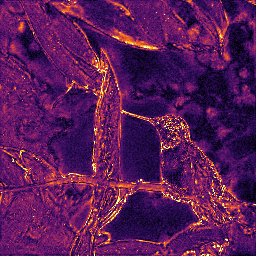} & \includegraphics[width=\ow]{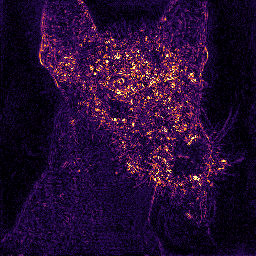} & \includegraphics[width=\ow]{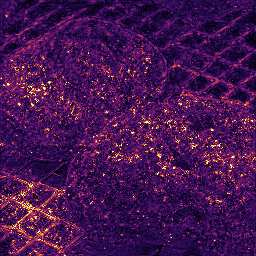} & \includegraphics[width=\ow]{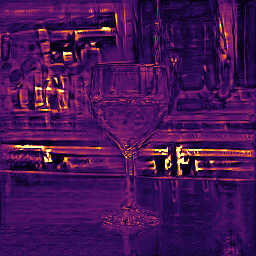} & \includegraphics[width=\ow]{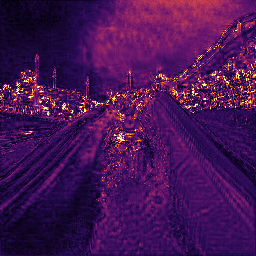} & \includegraphics[width=\ow]{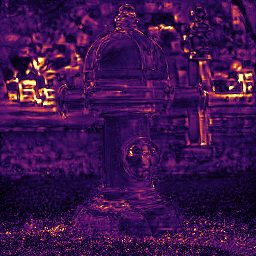} \\
\end{tabular}
\vspace{-4pt}
\caption{\textbf{Qualitative comparison on other generators.} (a) LightningDiT-XL, (b) REPA-E and (c) PixArt-$\alpha$. The same generated latents are decoded by $D_0$ and by \method. The bottom row shows their per-pixel absolute difference (brighter is larger).}
\label{fig:qual_more}
\end{figure*}

\figsuppunc

\noindent\textbf{Additional samples.} \Cref{fig:qual_more} extends the case study to LightningDiT, REPA-E and PixArt-$\alpha$, and \cref{fig:supp_unc} shows further DiT-XL/2 samples. Each pair decodes the same generated latent with $D_0$ and with the decoder trained on latents of the same generator, in the default setting of \cref{sec:exp:setup} (GAN with $\lambda{=}0.05$ on ImageNet and $\lambda{=}0.01$ on PixArt-$\alpha$); no post-processing is applied. Content, layout and colors are unchanged, while fur, feathers, food and fabric are rendered with more detail.

\phead{Strong guidance.} The last two samples of \cref{fig:supp_unc} are generated at guidance 4.0, a setting the decoder never sees in training. At this scale $D_0$ renders saturated and nearly flat surfaces; \method lowers the mean HSV saturation of these two images from $0.77$ to $0.59$ and from $0.88$ to $0.78$ and recovers the texture of the peel. This is the same correction as in the color statistics of \cref{tab:color}, and it agrees with the lower \fdclip{} at this guidance scale in \cref{tab:transfer}.

\phead{Other generators.} On LightningDiT and REPA-E, whose tokenizers are aligned with vision foundation models, the corrections are mainly in fine texture, such as feathers, fur and the crust of baked food. On PixArt-$\alpha$, \method also reduces its typical saturated colors (\cref{fig:supp_coco}).

\end{document}